\documentclass[10pt,journal,compsoc]{IEEEtran}

\ifCLASSOPTIONcompsoc

\usepackage{cite}
\usepackage{balance}
\usepackage{subcaption}
\usepackage{booktabs} 
\usepackage{graphicx}
\usepackage{verbatim}
\usepackage{amssymb,amsmath}

\usepackage{algorithmic}
\usepackage{slashbox}
\usepackage{breqn}
\usepackage{multirow} 
\usepackage{multicol}
\usepackage{rotating} 
\usepackage{epstopdf}
\usepackage{tikz}
\usepackage{courier}
\usepackage{microtype}
\usepackage[linesnumbered,ruled,vlined]{algorithm2e}
\usepackage{pifont}
\else
  \usepackage{cite}
\fi

\ifCLASSINFOpdf
\else

\fi

\begin{document}

\title{A Framework for Behavioral Biometric Authentication using Deep Metric Learning on Mobile Devices}

\author{Cong~Wang,
        Yanru~Xiao,
        Xing~Gao,
        Li~Li and
        Jun~Wang
\IEEEcompsocitemizethanks{\IEEEcompsocthanksitem C. Wang and Y. Xiao are with the Department
of Computer Science, Old Dominion University, Norfolk, VA, USA, E-mail: \{c1wang, yxiao002\}@odu.edu
\IEEEcompsocthanksitem X. Gao is with the Department of Computer Science, University of Delaware, Newark, DE, USA, Email: xgao@udel.edu
\IEEEcompsocthanksitem L. Li is with Shenzhen Institutes of Advanced Technology, Chinese Academy of Science, Shenzhen, China, Email: li.li@siat.ac.cn
\IEEEcompsocthanksitem J. Wang is with Futurewei Technologies, Santa Clara, CA, Email: jun.wang2@huawei.com
\IEEEcompsocthanksitem Correspondence to c1wang@odu.edu.}
}

\IEEEtitleabstractindextext{%
\begin{abstract}
Mobile authentication using behavioral biometrics has been an active area of research. Existing research relies on building machine learning classifiers to recognize an individual's unique patterns. However, these classifiers are not powerful enough to learn the discriminative features. When implemented on the mobile devices, they face new challenges from the behavioral dynamics, data privacy and side-channel leaks. To address these challenges, we present a new framework to incorporate training on battery-powered mobile devices, so private data never leaves the device and training can be flexibly scheduled to adapt the behavioral patterns at runtime. We re-formulate the classification problem into deep metric learning to improve the discriminative power and design an effective countermeasure to thwart side-channel leaks by embedding a noise signature in the sensing signals without sacrificing too much usability. The experiments demonstrate authentication accuracy over 95\% on three public datasets, a sheer 15\% gain from multi-class classification with less data and robustness against brute-force and side-channel attacks with 99\% and 90\% success, respectively. We show the feasibility of training with mobile CPUs, where training 100 epochs takes less than 10 mins and can be boosted $3$-$5$ times with feature transfer. Finally, we profile memory, energy and computational overhead. Our results indicate that training consumes lower energy than watching videos and slightly higher energy than playing games.
\end{abstract}

\begin{IEEEkeywords}
behavioral authentication, on-device deep learning, privacy preservation, deep metric learning.
\end{IEEEkeywords}}

\maketitle

\IEEEdisplaynontitleabstractindextext

%
\IEEEpeerreviewmaketitle


\vspace{-0.05in}
\IEEEraisesectionheading{\section{Introduction}\label{sec:introduction}}

Smartphone has become an indispensable part of our lives. Being a rich mine of personal and corporate data, it is usually sought by cybercriminals as a gateway to key information. The cost of losing one is beyond the replacement of hardware, since data found on the device is usually private and sensitive. As a result, over 65\% of users have enabled authentication on their devices \cite{soups}. The security-usability conflict has been a persistent design challenge of mobile authentication. From the early methods of passcode, swipe pattern, to the recent advance of fingerprints and FaceID, authentication still requires deliberate attention from the user. On the other hand, behavioral biometrics~\cite{gait1,gait2,gait,gait-ndss,touch,ipccc,zhou1,caba,ndss_key,gao,breathprint} provide implicit channels to identify the user, as they reflect the internal characteristics of a user, and are difficult to replicate. These have made behavioral biometrics a promising (second-factor) authentication to resolve the fundamental usability issues.

Unfortunately, the identification accuracy is still far from satisfaction in the production environments. The behavioral dynamics evolve due to a complex combination of external factors such as sickness, injury and emotion, thus intensifying the intra-class variations and hamper classification. Existing research built around machine learning (ML) leverages statistical properties to recognize users \cite{gait,gait1,gait2,gait-ndss,touch,ipccc,zhou1,caba,ndss_key,gao,breathprint}. They barely meet practical requirements since hand-crafted statistical features have low discriminative power. Their implementations further ignore important aspects of where and how the model should be trained: these always have profound security and privacy implications. For example, they loosely aggregate user data for training without security precautions. Users may be unwilling to upload private data. On the system level, the read access to motion sensors are not restrictive in the OS~\cite{android_permission,ndss_track,grophone,password,slogger,deepmag}, a strong attacker can exploit these side channels to gather motion data. As the same data is also used to train the classifier, the attacker can launch a replay attack by programming an apparatus to generate the identical pattern as the sniffed data~\cite{lego-robot}. The existing research has yet to consider these issues in the context of mobile authentication.

Meanwhile, propelled by the latest advance in mobile processors, smartphone becomes an ideal platform to execute ML at the data source. The recent efforts mainly improve \emph{inference} from a pre-trained deep neural network model~\cite{lowpgemm,mobilenetv2,song,distillation}. Yet, inference from a static model still has a significant gap from being cognizant, since ML relies on the basics that the test samples are independently and identically drawn from the same distribution of training. Deep classifiers are not good at extrapolation when data comes from a different distribution, but it is quite common in mobile applications. This requires the model to adapt to the new distribution, e.g., re-training or finetuning the model to tackle behavioral dynamics. Therefore, being truly intelligent should bring \emph{training} back into the loop.

A natural solution is to securely offload training to the cloud. One can host a secure enclave for each user and securely schedule re-training on-demand~\cite{sgx}. However, it also comes with some limitations: 1) \emph{Network connectivity}. Network connection is necessary to upload the new data and receive the updated model. Users may be hesitant when their LTE data plan is metered. 2) \emph{Scalability and Service Cost}. Such centralized approach is not scalable to a large population of users as each user requests dedicated resources from cloud servers. Either the end user or the service provider has to pay for the service, e.g., the Amazon's containerized application at \$0.04 per CPU/hr~\cite{aws_pricing} would accumulate to a large amount if constant training is required. Other privacy-preserving techniques are also available~\cite{dp_deep,cryptonets,iotdi}, but they often come with a nontrivial accuracy loss.

Can we just incorporate training on mobile devices? What are the gaps between mobile and ML that limit the algorithms to perform as they should? Instead of the foreseeable challenges of computation and power in the system architecture, this paper focuses on application-level challenges of \emph{privacy} and \emph{accuracy} to build a comprehensive framework for behavioral authentications. To achieve high accuracy in authentication, we re-formulate the classification problem into deep metric learning~\cite{siamese-lecun,koch2015siamese} to learn a distance metric between the similar and dissimilar samples, so that the features have high discriminative power. Since metric learning learns from pairwise inputs, we develop a new sampling technique that achieves data balance within the memory constraint. After the model is trained, to enhance the reliability of decisions in vibrant mobile environments, we develop a space-time decision fusion algorithm that generates a decision with high confidence level from multiple inference. The results can be fed back to schedule training hence close the loop of learning. Although on-device learning preserves data privacy by default, side-channel leaks from motion sensors in the OS leaves the door open for attackers to spoof authentication via replay attacks~\cite{lego-robot}. We build a new defense mechanism into the framework to recognize and reject fraudulent samples sniffed from the motion sensors. It is achieved by embedding a noise fingerprint inside the sensing signal and supervising the neural network to distinguish between genuine samples from the ones with noise perturbation. We also explore using feature transfer to speed up training convergence on mobile, while securing all the intermediate activations/model parameters. The main contributions are summarized below.

\begin{itemize}
  \item We develop a framework for behavioral authentication that the entire loop of training and inference are executed on device. The implementation on Android demonstrates that training is not only feasible but also quite fast with feature transfer (within 5s/epoch on Huawei Mate10 for $400$ samples).
  \item We enhance the discriminative power of the classifier through deep metric learning. Our experiments demonstrate 10-15\% improvement of authentication accuracy on different datasets and achieve an accuracy of $0.94$ on a large dataset with $153$ participants. We also analyze and compare various types of data representations and loss objectives thoroughly.
  \item We successfully mitigate potential side-channel leaks and reduce attack success ratio below 10\% while preserving the usability of benign applications.
  \item We evaluate the framework extensively on three public datasets and profile learning performance and cost on various smartphone models. To the best of our knowledge, this is the first work that implements both training and inference, and addresses the associated challenges for behavioral authentication on battery-powered mobile devices.
\end{itemize}

The rest of the paper is organized as follows. Section \ref{background:sec} studies discusses the background and related literatures. Section \ref{model:sec} and Section \ref{design:sec} present the system model and design. Section \ref{evaluation:sec} evaluates the framework with necessary discussions in Section \ref{discussion:sec} and Section \ref{conclusion:sec} concludes this work.

\vspace{-0.1in}
\section{Background and Related Works}   \label{background:sec}

\vspace{-0.05in}
\subsection{Behavioral Authentication}
Commodity mobile devices usually feature an array of sensors that capture the information of acceleration, orientation, angular velocity, magnetic field, etc. These data always implies the behaviorial biometrics of the user, as gait~\cite{gait,gait1,gait2,gait-ndss,segauth}, touch patterns~\cite{touch}, keystroke dynamics~\cite{ndss_key}, micro-movement \cite{sp_workshop,ipccc} or their combinations~\cite{zhou1,caba}, eye movement~\cite{gao} and even breath~\cite{breathprint} are proven to be successful in differentiating human subjects. A system process can run continuously in the background for implicit authentication with no deliberate attention from the user~\cite{itus}, which makes behavioral biometrics an ideal \emph{second factor} for authentication. For example, in~\cite{touch}, four tapping features are analyzed when users type five different PINs. Micro-movements of the phone are considered along with touch-based~\cite{ipccc} and signature-based~\cite{sp_workshop} features for authentication. Multi-modalities from hand movement, orientation, and grasp features are considered in~\cite{zhou1}. An ensemble of biomedical signals is gathered by medical sensors for continuous authentication in~\cite{caba}. Audio signatures from sniff, normal and deep breathing are captured via the microphone to identify users~\cite{breathprint}. This paper adopts \emph{gait} as a representative among behavioral biometrics. Clinic research has suggested gait as a strong indicator to distinguish between human subjects~\cite{gait,gait1,gait2,gait-ndss,segauth}. Statistics of correlation, fourier coefficients, histogram are analyzed in~\cite{gait1} and more statistical features are considered in~\cite{gait2}. In~\cite{gait}, an algorithm is proposed to detect gait cycles and compute similarity score with a heuristic algorithm. In~\cite{gait-ndss} and~\cite{segauth}, time series are segmented for similarity measurement using dynamic time warping, that relies on the alignment of gait cycles.

The research of behavioral identification has been on the side of exploring different modalities to identify the user. Statistical features such as average max, min, median acceleration, absolute distance, standard deviation, histogram, periodicity, Fourier coefficients or Discrete Cosine Transform coefficients~\cite{gait,gait1,gait2,gait-ndss} are typically extracted to train a classifier using a dataset collected in confined environments with minimum interference. In practice, outliers and abrupt changes could easily mislead those statistical classifiers with less discriminative power. To this end, recent studies start to adopt deep neural networks for ``automated'' feature extraction~\cite{idnet,motion}. Though promising results are shown, they still leave a considerable gap from the high-accuracy requirement in production environments. Furthermore, a common oversight from the previous works is \emph{where} and \emph{how} the model should be trained, in the shadow of potential threats from privacy leakage on cloud servers to side channel exploits on mobile devices. In this paper, we develop a comprehensive framework to incorporate the entire loop from data generation, computation and decision making on mobile devices, and design a mechanism to defend against side-channel replay attacks.

\vspace{-0.1in}
\subsection{Privacy Preserving and On-Device Learning}

Privacy preservation has been extensively studied recently in the machine learning community. The research takes three main directions. The first is the algorithmic direction such as differential privacy~\cite{dp_deep} and data projection~\cite{iotdi}. Differential privacy introduces noise into the training process so adversaries cannot detect the presence or absence of a user~\cite{dp_deep}. Autoencoder is utilized to transform sensitive features into a latent space for non-sensitive inference~\cite{iotdi}. Since these schemes are always subject to the privacy-utility conflict, it is hard to provide rigorous security guarantees and an accuracy loss has to be paid. The second direction is homomorphic encryption that allows curious third parties to perform meaningful computations on encrypted data~\cite{cryptonets,ijcai}. CryptoNets use fully homomorphic encryption to encrypt the data from the client and receive an encrypted prediction from the cloud. A square activation function is adopted to bridge the gap between the cryptographical and neural operations, which is only suitable for inference computation due to the accumulation of error in training. The work of~\cite{ijcai} designs a two-party protocol to protect both the data and the model using a partially homomorphic encryption. Since homomorphic encryption and decryption are computationally intensive, the computation, energy and communication overhead is prohibitive for mobile devices.

\emph{On-device Learning.} The third approach is to implement computations directly at the data source, so private information never leaves the device. The existing works either develop new variants of applications such as activity recognition~\cite{rfid} and mobile vision~\cite{zhang,lane}, or optimize inference execution on mobile devices using reduced precision and weight pruning~\cite{lowpgemm,song}. These techniques aim to reduce the run-time redundancy of the neural networks. A direct method is to round the original model parameters in 32-bit floating point to 8-bit integer, so 75\% size can be saved~\cite{lowpgemm}. Another approach is to prune the connections with near-zero weights after training~\cite{song}. A large and complex teacher network can be also used to train a small student network for comparable results, thus distilling the knowledge to run the small network on mobile devices~\cite{distillation}. These methods are effective for running inference on mobile devices. Nevertheless, they left training out of the loop and are not able to timely engage the dynamics in mobile applications.


Training needs to store all the intermediate results (feature maps, convolution outputs) in the memory for backpropagation. This demands at least twice memory usage compared to inference~\cite{super-neurons,layup}. For inference, as long as the layers are not referenced any more, its memory can be released and re-used to reduce the peak memory footprint~\cite{cash}. For training, the intermediate results are persistent in memory until being consumed during the backpropagation, thereby rendering less chance to optimize. The existing architecture and software stack target at mobile workloads that are bursty in nature (user interactions) such as web browsing, texting, etc. The emergence of sustained, highly paralleled workloads have not been fully explored so far. Whether the memory capacity, power management policy and multi-core CPUs are capable to handle neural computation and, meet the application-level requirements are still unknown. This work fills such gap to explore the challenges of training by building the entire framework of behavioral authentication on mobile devices.

\emph{Side-channel Attacks on Motion Sensors.} Android, one of the most popular mobile OS, only asks for permissions to access sensitive data (such as contacts, messages, camera, GPS, etc)~\cite{android_permission}. However, the read access from other sensors is not restricted. This has opened the door for attackers to launch various kinds of motion-based side-channel inference attacks. For example, early works have shown that the accelerometer can be exploited to extract text inputs from the touchscreen~\cite{password}. The works later exploit other sensing vectors to recognize speech from gyroscope~\cite{grophone}, sniff app usage from magnetometer~\cite{deepmag} or utilize manufacture nuances to fingerprint users~\cite{ndss_track}. Such side-channel vulnerability has direct impact on the implementation of authentication, an essential risk that has been overlooked in the behavioral authentication community~\cite{gait,gait1,gait2,gait-ndss,segauth,touch,ndss_key,sp_workshop,ipccc,zhou1,caba}. Powerful attackers can launch replay attacks from sniffed sensing data, e.g., programming a robot to generate the same accelerometer reading to bypass gait authentication~\cite{lego-robot}. Common defense includes injecting random noise or reduce sampling rates to obfuscate the sensor data~\cite{ndss_track,slogger}. Their effectiveness depends on how much the data is ``blurred'' from the original signal, which adversely impacts the usability of the normal applications. This paper integrates a joint design of authentication with noise injection, by utilizing the high representational power of neural networks, so that the noise level required for protection can be reduced significantly.

\vspace{-0.1in}
\subsection{Deep Metric Learning}
Loss function is important for achieving high accuracy of deep models. \emph{Softmax} loss is the most popular one in classification tasks, which consists of the output feature vector (the penultimate layer), the softmax function and the cross-entropy loss. The softmax function produces a probability distribution over the class labels. While the correct class labels are represented by one-hot encodings, the cross-entropy loss are trained to make the output feature vector as close as possible to the one-hot vector. Softmax is widely adopted in the state-of-the-art CNN designs~\cite{alexnet,vgg,resnet} because of its mathematical tractability and clear probabilistic interpretation. Unfortunately, it only encourages the separation of features, but the learned features are not discriminative enough~\cite{icml16,eccv16}. A quick fix is to increase the amount of data or the depth of the network. However, gathering more data also involves nontrivial efforts to cleanse, augment and label the new data; increasing the network depth could easily violate the peak memory bound on embedded devices. As a result, softmax still falls short in many large-scale identification tasks~\cite{facenet} as its accuracy cannot meet the strict requirements in practice, particularly, for security-critical applications such as authentication.

Instead of mapping feature vectors in a closed set of class labels, deep metric learning learns the relative distances between samples via a pairwise similarity loss function and transforms features into a space that can be measured by, e.g., the Euclidean distance~\cite{siamese-lecun,koch2015siamese}. In softmax function, the gradient is computed for each sample with the correct class label, whereas the gradients in metric learning depend on the pairwise correlations between samples. Given a similarity metric, similar samples are projected into neighborhood locations in the metric space and dissimilar samples are pulled apart. It has achieved higher accuracy than softmax in different tasks of facial recognition~\cite{facenet}, object classification~\cite{obj_reco} and person re-identification~\cite{re-id}. The design of the loss function determines the discriminative power and the corresponding sampling efforts. For instance, the Siamese network~\cite{siamese-lecun,koch2015siamese} adopts the contrastive loss. The Triplet network introduces an additional anchor point and generates a ranking of the closest images to an input~\cite{iclr_triplet,facenet}. However, triplets also expand the pairwise sampling effort from $\mathcal{O}(n^2)$ to $\mathcal{O}(n^3)$. The computational cost also grows linearly with the number of neural network branches. Further extensions have adopted quadruplets~\cite{quadruplet} at much higher sampling ($\mathcal{O}(n^4)$) and computational cost. In this paper, we focus on the Siamese Network given its moderate space/computational complexity and explore the design space on resource-constrained embedded devices under limited data, storage and memory space to bring the entire loop of authentication on-device.

\section{System Architecture and Model} \label{model:sec}
\begin{figure*}
\centering
\vspace{-0.08in}
\hspace*{-0.5cm}
\includegraphics[width=5.8in]{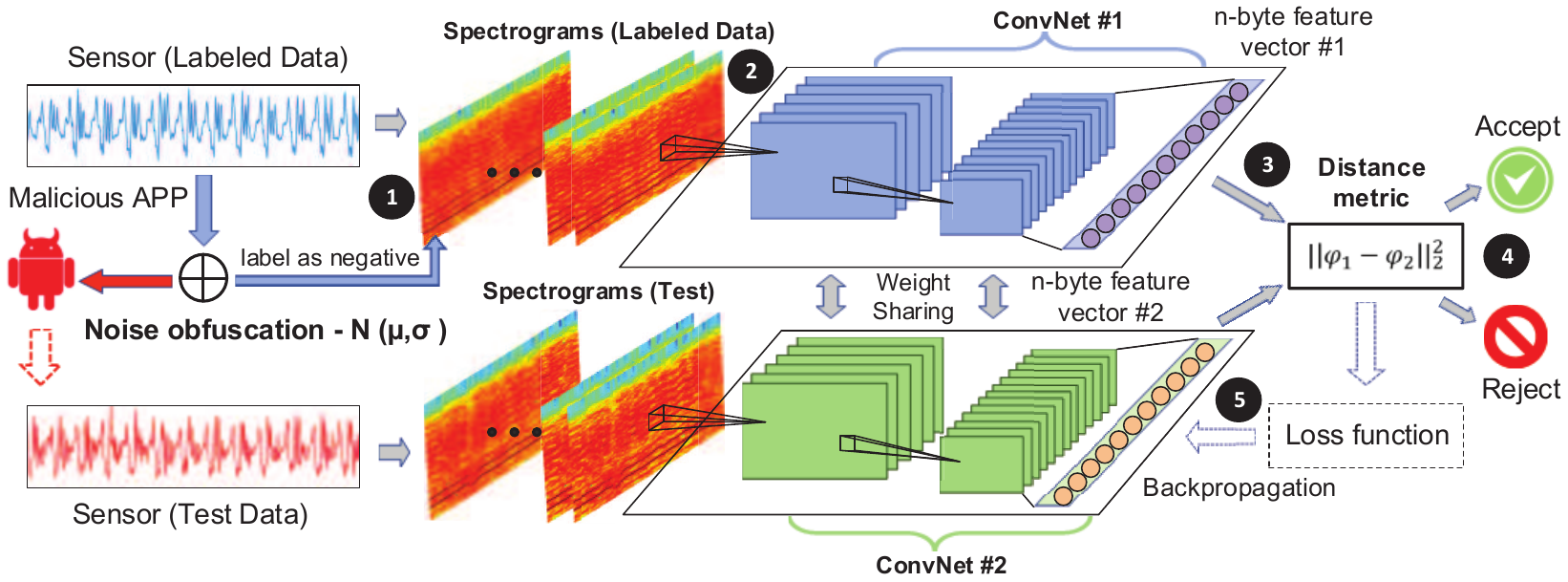}
\vspace{-0.05in}
\caption{System architecture on mobile devices: \ding{182} it takes raw sensor inputs, transforms them into mid-level representations (spectrograms~\cite{spectrogram}); \ding{183} processes the representations with the neural network; \ding{184} computes a distance metric from the feature vectors; \ding{185} generates a decision; \ding{186} backpropagates the error if training is scheduled.}
\label{overview}
\vspace*{-0.2in}
\end{figure*}

In this section, we describe the system architecture and threat model for deep behavioral authentication. The proposed framework is depicted in Fig.~\ref{overview}.

\noindent \textbf{Threat Model.} The authentication module reads authentic sensor data from the standard API (e.g., \texttt{SensorManager} in Android). We assume that sensor data is trustworthy as its integrity can be protected by hardware security extension technologies such as ARM TrustZone~\cite{arm_trustzone}. Meanwhile, sensor data is available to all applications, including malware, since modern OS like Android or iOS does not set restrictions to accessing sensors. By exploiting sensor data (e.g.,  accelerometer, gyroscope, magnetometer), numerous side-channel attacks~\cite{password,grophone,ndss_track,deepmag}  have been reported to sniff PIN/swipe patterns, password, app usage or even speech.

The main performance metric is the authentication accuracy, measured by whether the system can defend against spoofing attacks and recognize its owner. Specifically, we consider two types of attackers: \emph{passive} and \emph{active}. \textit{Passive} attackers~\cite{ndss_key} use their own data or samples from a large database to spoof the authentication system. This case is common since a random attacker may obtain a device lost by the victim.  Without any prior knowledge on the behavioral pattern, the passive attacker can only retrieve data from a large public database and launch brute-force attacks to unlock the device. \textit{Active} attackers can directly and stealthily collect sensory data from the user's smartphone. This could be achieved by tricking the user into installing a third-party app.  We further assume the powerful active attacker can generate the exact same gait pattern as the sniffed sensory data by programming an apparatus such as a robot~\cite{lego-robot}.

One effective countermeasure to side-channel attacks is sensor data obfuscation~\cite{ndss_track}, which injects random noise to user-level apps so that malware cannot recognize sensitive operations. The noise level should balance the usability of benign apps while obfuscating malicious ones. Thus, to mitigate side-channel attacks, the obfuscation technique is employed in the target smartphone by wrapping the \texttt{SensorManager} API. For instance, Slogger~\cite{slogger} can inject noise into various sensor outputs. The noise is transparent to the authentication module. While applications (including malware) can only obtain obfuscated data through the wrapped interface, they can apply various denoising techniques to restore the original data. Since our design is centered around on-device implementation, the cloud only plays a secondary role to provide samples from the negative classes. We assume cloud providers are honest but curious: they follow protocols but are free to use what they see to learn private information from users.


\vspace{-0.05in}
\section{Framework Design}   \label{design:sec}
This section presents the main design of the framework, including the pair sampling, learning technique, decision-making mechanism and opportunities to speed up the computation.

\vspace{-0.1in}
\subsection{Memory-efficient Pair Sampling}
This section presents a new sampling technique to construct balanced pairs under the memory limit. The Siamese Network takes pairwise inputs: a) positive pairs, that pair similar samples from the same class; b) negative pairs, that pair dissimilar samples from different classes. The positive samples are collected on the mobile device during the bootstrapping phase and the negative samples are supplied by the cloud as pre-loaded into the storage. Training takes batched input in memory sampled from flash storage. To avoid the latency accessing the storage, the framework maintains a pool of sampled pairs in memory. This makes sampling crucial because of data imbalance and increased memory footprint.

In authentication, the number of negative samples is much larger than the positive ones. For $n$ negative classes with $s$ samples of each class and $r$ positive samples collected at the mobile user, there are $r^2$ positive pairs and $n \cdot s \cdot r$ negative pairs. With a large number of negative classes, $n \cdot s$ is much larger than $r$. Hence, the total number of negative pairs is much larger than the positive ones. Since loading all the negative pairs into memory may lead to memory error, the goal is to keep a random subset within memory limits.

We develop a balanced reservoir sampling algorithm based on~\cite{vitter1985random}. A buffer size of $2R$ is found from hardware configuration or test (half for positive and half for negative pairs). The size determines a trade-off between memory usage and variety of negative records. Small $R$ could lead to severe overfitting and large $R$ risks of having memory error. To maximize coverage, we set $R=r^2$ so all positive samples are utilized for training and make sure that the total size of $2R$ is within the memory capacity. The algorithm continuously adds record into the reservoir till the $(T+1)$-th record, $T = R$. If $T > R$, a random pair in the reservoir is replaced with probability $\frac{R}{T}$ or the reservoir is kept the same with probability $1-\frac{R}{T}$. After the sequential pass through all the records, the buffer forms a random set from the pool of negative samples.

\begin{algorithm}[tb]
\footnotesize
  \caption{Memory-efficient Sampling}
  \label{sampling_algo}
  \textbf{Input}: $r^2$ positive and $n \cdot r \cdot s$ negative pairs, memory bound $2R$.\\
  \textbf{Output}: a balanced set of samples of size $2R$.\\
  Set of all negative pairs $\mathcal{N}$, $R=r^2$, $|\mathcal{N}| = n \cdot r \cdot s$. \\
  \For{$T \leftarrow 1,\cdots,R$}
  {
     $\mathcal{R} \leftarrow \mathcal{R} + (i \in \mathcal{N})$.
  }
  \For{$T \leftarrow R+1,\cdots,n r s$}
  {
      \If{probability $p > \frac{R}{T}$}
      {
        $\mathcal{R} \leftarrow \mathcal{R} - (i \in \mathcal{R}) + (i \in \mathcal{N})$.
      }
      \Else
      {
        $\mathcal{R} \leftarrow \mathcal{R}$.
      }
  }
\end{algorithm}

\vspace{-0.1in}
\subsection{Feature Embedding from Siamese Network}

The existing work relies on the \emph{softmax} loss with less discriminative power~\cite{idnet}. This section describes a new set of feature embeddings learned from the Siamese Network. As shown in Fig. \ref{overview}, the Siamese Network incorporates two identical branches of convolutional neural networks with shared model weights. First, the 1D sensing signal is pre-processed (discussed in Section \ref{sec:spectrogram}), converted into a 2D representation and made into positive and negative pairs following Algorithm \ref{sampling_algo}. The Siamese Network takes a pair of inputs to feed into the two identical branches and yields two feature embeddings $\varphi_1, \varphi_2$. A distance metric function $f(\varphi_1, \varphi_2)$ is applied to compute the distance between them, e.g., the $l_2$ distance $f(\varphi_1, \varphi_2)=\lVert \varphi_1 - \varphi_2 \rVert_2$. The distance is used to train a \emph{contrastive loss}~\cite{siamese-lecun} and map feature vectors to a space in which similar samples have closer distance whereas dissimilar samples are far apart (separated by a margin). For a pair $i,j$ of dataset $\mathcal{D}$, the contrastive loss function is defined as,
\begin{eqnarray}
\vspace{-0.1in}
\small
&&\mathcal{L}_c = \sum\nolimits_{i,j \in \mathcal{D}} y(\varphi_1^{(i)}, \varphi_2^{(j)}) f(\varphi_1^{(i)}, \varphi_2^{(j)})^2+ \nonumber \\
&&(1-y(\varphi_1^{(i)}, \varphi_2^{(j)})) \max(m - f(\varphi_1^{(i)}, \varphi_2^{(j)}),0)^2, \label{eq_contrastive}
\end{eqnarray}
in which label $y(\varphi_1^{(i)}, \varphi_2^{(j)})=0$ for dissimilar pairs and $y(\varphi_1^{(i)}, \varphi_2^{(j)})=1$ for similar pairs. $m$ is the margin that separates the dissimilar samples. If the pair is similar (positive), the loss is $f(\varphi_1^{(i)}, \varphi_2^{(j)})^2$; if the pair is dissimilar (negative), the loss is $\max(m-f(\varphi_1, \varphi_2))^2$. When $f(\varphi_1, \varphi_2)>m$, the loss is zero, i.e., dissimilar pair with distance larger than the margin has zero loss. It means that the loss would not encourage further separation when the distance between a dissimilar pair is already larger than the margin.

\textbf{Joint Loss.} A slightly different loss function $\mathcal{L}_s$ is proposed in~\cite{koch2015siamese}, that maps dissimilarity into a probability prediction with sigmoid activation so the network can be trained with cross-entropy loss. The advantage is that no margin needs to be pre-determined,
\begin{multline}
\small
\vspace{-0.08in}
\tiny
\mathcal{L}_s = \sum\nolimits_{i,j \in \mathcal{D}} y(\varphi_1^{(i)}, \varphi_2^{(j)}) \log p(\varphi_1^{(i)}, \varphi_2^{(j)})+
\vspace{-0.05in}
\\(1-y(\varphi_1^{(i)}, \varphi_2^{(j)})) \log (1- p(\varphi_1^{(i)}, \varphi_2^{(j)})).
\vspace{-0.08in}
\label{eq_cross_entropy}
\end{multline}
The construction of the loss function plays an important role to capture similarity in different applications. To capitalize from the potential advantages of distance and probabilistic metrics, we combine them into a new joint loss function. The goal is to minimize the total loss $\mathcal{L}_t$ with a balancing parameter $\alpha$,
\begin{equation}
\vspace{-0.03in}
\small
\mathcal{L}_t = \mathcal{L}_c + \gamma \mathcal{L}_s, \label{eq_joint_loss}
\vspace{-0.03in}
\end{equation}
where $\mathcal{L}_s$ is the cross-entropy loss in Eq. (\ref{eq_joint_loss}) and $\mathcal{L}_c$ is the contrastive loss in Eq. (\ref{eq_contrastive}). $\gamma$ is a scaling parameter used for balancing the two functions. The feature embeddings can be directly measured in terms of a distance metric (i.e., $l_2$ distance in this paper), and these different loss formulations are evaluated in Section \ref{evaluation:sec}.




\vspace{-0.1in}
\subsection{Decision Fusion and Feedback}   \label{sec:sprt}
After the model is trained, the inference module takes input from sensors and outputs a classification decision. The decision based on a single shot of inference is not reliable because interference, outliers, and behavioral instability persist at run-time. The goal is to reach a high confidence within minimum observation time. We build an algorithm on top of the inference module to fuse multiple inferences across the spatial and temporal dimensions. We leverage the training samples as the ground truth and assess whether the input data is similar or dissimilar to the training samples, i.e., pair the input (testing data) with training samples. Since paring with one training sample is not sufficiently representative, for input $x_i$ at time $i$, $x_i$ is paired with $k$ samples randomly selected from the training set on mobile ($k>1$). The average distance from $x_i$ to the $k$ samples are computed as $d_i = \sum_{j=1}^k d(x_j, x_i)/k$, in which $d(x_j, x_i)$ is the distance between input $x_i$ and training sample $x_j$.


Not only could the selection of training samples have imperfections, the incoming data may also have disturbances. After the spatial evaluation, we progress along the time dimension to fuse multiple decisions $\{y_1, y_2, \cdots, y_n\}$. After the $i$-th evaluation, it either decides to \emph{accept} ($H_0$), \emph{reject} ($H_1$) or \emph{continue} to observe $y_{n+1}$. The module defines two kinds of errors: false negative $\alpha$ and false positive $\beta$. The objective is to minimize the expected time of evaluation and satisfy the error constraints, which is formulated as Sequential Probability Ratio Test (SPRT)~\cite{wald}. SPRT progresses by assessing a likelihood ratio $\lambda_n$ for the $n$-th observation,
\begin{equation}
\small
\lambda_n = \frac{p(y_1,\cdots,y_n | H_1)}{p(y_1,\cdots,y_n |H_0)} = \prod_{i=1}^{n} \frac{p(y_i | H_1)}{p(y_i |H_0)}.  \label{likelihood}
\end{equation}
The second equality holds because samples are independently randomly drawn. We extend SPRT for the distance metric (contrastive loss). Pairs with distance less than the margin threshold are considered as similar; otherwise, they are dissimilar. To convert the distance into probabilistic representation, we approximate it with a normal distribution but in an upside-down bell shape, which has been used for risk assessment to impose a loss on off-target products in manufacturing~\cite{upside-down}. We adopt the same analogy here since distances that are close to 0 or margin $m$ have a high probability of being similar or dissimilar respectively. The margin threshold is typically set to $\frac{m}{2}$ for a balanced set of positive and negative pairs. The distance around $\frac{m}{2}$ means the classifier is unsure about the pairs so a lower probability is given.
\begin{equation}
\small
p(d_i| \mu , \sigma^2) = 1 - \phi(\frac{d_i - \mu}{\sigma^2}). \label{proability}
\end{equation}
The mean and variance can be obtained from the learned distance of the training data. Combining \eqref{likelihood} and \eqref{proability}, the ratio is,
\begin{equation}
\small
\frac{p(y_i=0| H_1)}{p(y_i=0| H_0)} = \phi(\frac{d_i - \mu}{\sigma^2})/\big(1-\phi(\frac{d_i - \mu}{\sigma^2})\big)
\end{equation}
\begin{equation}
\small
\frac{p(y_i=1| H_1)}{p(y_i=1| H_0)} = \big(1-\phi(\frac{d_i - \mu}{\sigma^2})\big)/\phi(\frac{d_i - \mu}{\sigma^2})
\end{equation}
According to~\cite{wald}, the decision strategy is,
\begin{equation}
\small
S_n^\ast = \left\{
             \begin{array}{lr}
             H_0, \lambda_n \leq B\\
             H_1, \lambda_n \geq A\\
             continue, B < \lambda_n < A
             \end{array} \label{sprt_eq}
\right.
\end{equation}
We set the two thresholds $A$ and $B$ suggested by~\cite{wald}, $A = (1-\beta)/\alpha$, $B=\beta/(1-\alpha)$. The sequence moves within the open interval $(B,A)$ till a decision is made. Intuitively, if consecutive decisions of acceptance are made, the likelihood ratio shrinks multiplicatively. Any rejection along the way would drive the ratio to an opposite direction towards the upper threshold until a threshold is met. The decision of $S_n^\ast$ is examined closely to schedule training.

\textbf{Feedback.} We examine the testing accuracy as a feedback to schedule model re-training and adapt variations. In authentication, if the decision outputs a false negative, the screen is mistakenly locked by the (second-factor) behavioral authentication, but the user later logins with her face or fingerprint (that verifies the decision is indeed a false negative). If such situations exceed a certain number, it indicates that the user's behavior may have undergone a substantial change and training is scheduled with a mix from the new data. Incorporating training on mobile could immediately respond to these deviations thereby closing the loop of learning on mobile devices. The scheme is summarized in Algorithm \ref{sprt_algo} and evaluated in Section~\ref{sec:miscellaneous}.

\begin{algorithm}[tb]
\footnotesize
  \caption{Decision Fusion and Feedback}
  \label{sprt_algo}
  \textbf{Input}: Testing pairs $(x_j, x_i)$, $1 \leq j \leq k$. $k$ pairs randomly drawn from training set. False negative $\alpha$ and false positive $\beta$, threshold $A=(1-\beta)/\alpha$, $B=\beta/(1-\alpha)$.\\
  \textbf{Output}: Decision $S_n^\ast$ and training schedules. \\
  Initialize false negative counter $c \leftarrow 0$, and threshold $T$.  \\
  \While{$c<T$}
  {
      $n \leftarrow 0$\\
      \While{$B < \lambda_n < A$}
      {
          $\overline{d_i} \leftarrow \sum_{j=1}^k d(x_j, x_i)/k, p(d_i| \mu , \sigma^2) \leftarrow 1 - \phi(\frac{d_i - \mu}{\sigma^2})$. \\
          $\lambda_n \leftarrow \prod_{i=1}^{n} \frac{p(d_i | H_1)}{p(d_i |H_0)}$. \\
          \If{$\lambda_n \geq B$}
          {
              $S_n^\ast \leftarrow 1$ and \textbf{Break}.
          }
          \If{$\lambda_n \leq A$}
          {
              $S_n^\ast \leftarrow 0$ and \textbf{Break}.
          }
          $n \leftarrow n+1$
      }
      Output optimal decision $S_n^\ast$.\\
  \If{Given true label $H_0$, $S_n^\ast = H_1$.}
  {
    $c \leftarrow c+1$
  }
  }
  Schedule training of $\mathcal{M}_t$ with new data $\mathcal{D}_t$.
\end{algorithm}

\begin{figure*}[!t]
\centering
\hspace*{-0.3in}
\begin{subfigure}[b]{0.3\textwidth}
                \includegraphics[width=1.1\textwidth]{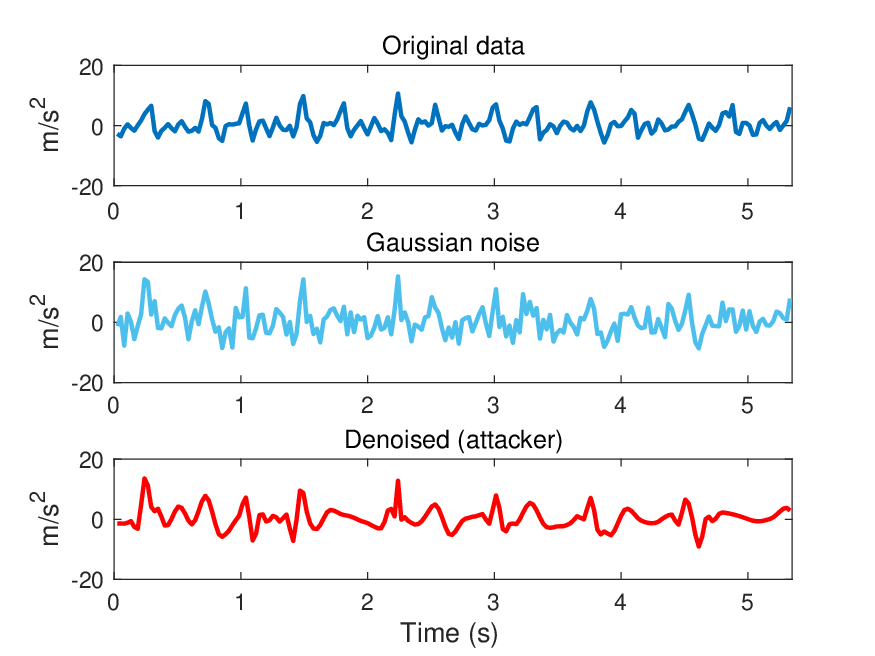}
                \vspace{-0.2in}
                \caption{}
\end{subfigure}
\hspace*{0.05in}
\begin{subfigure}[b]{0.3\textwidth}
                \includegraphics[width=1.1\textwidth]{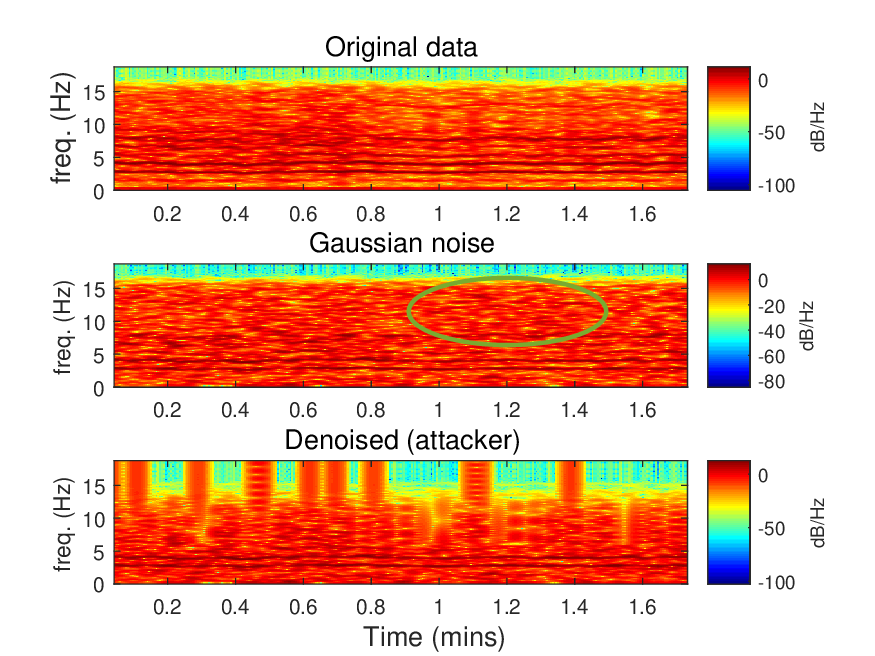}
                \vspace{-0.2in}
                \caption{}
\end{subfigure}
\hspace*{0.05in}
\begin{subfigure}[b]{0.29\textwidth}
                \includegraphics[width=1.1\textwidth]{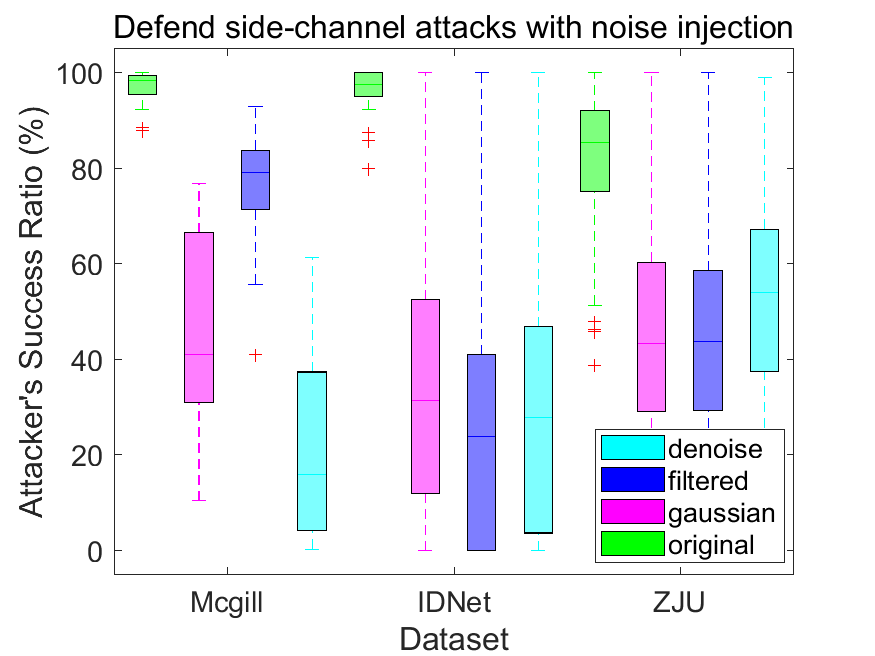}
                \vspace{-0.2in}
                \caption{}
\end{subfigure}%
\vspace*{-0.11in}
\caption{Preliminary assessment of simple noise injection (a) raw sensing data with/without noise; (b) relevant spectrograms; (c) attacker's success ratio.}
\label{side_channel_prelim_fig}
\vspace*{-0.2in}
\end{figure*}

\vspace{-0.1in}
\subsection{Defend Side-channel Leaks and Active Attacks}   \label{sec:side_channel_attack}


Although well-trained neural networks could achieve high accuracy, the sensitive data might be also leaked via side channels. Skilled attackers can trick the user into downloading an app that stealthily captures the motion data, and then replay them to gain the access via programming an apparatus~\cite{lego-robot}. A typical countermeasure is to obfuscate the sensor output by injecting random noise~\cite{ndss_track}. However, our experiments show that simple noise injection fails to fully prevent the attack. Specifically, we obfuscate the data with a \textit{zero-mean} gaussian noise, whose standard deviation is set to equal the original signal over a moving window. As circled in the middle picture of Fig.~\ref{side_channel_prelim_fig}(b), the obfuscation introduces a few new energy components at higher frequencies in the spectrograms, which enable the authentication module to recognize attackers to some extent.
Fig.~\ref{side_channel_prelim_fig}(c) shows the attacker's success ratio. Without any protection, the attacker can easily accomplish 80-100\% success rate (the true positive rate). With noise injected, the success rate drops to an average of 50\%. However, it is still not sufficiently secure. In extreme cases, some individuals are still subject to 100\% attack success rate even when noise is injected.

One reason that simple noise injection does not work here is that neural network is robust to random noise. It extracts a meaningful combination of features towards a minimization of the loss objective, and serves as an information bottleneck that finds a compressed mapping of input that preserves maximally possible information of the output~\cite{bottleneck}. Thus, redundant information including small noise and interference, which does not interfere with the main structure, is thrown away. As a result, the attackers can still succeed. Obviously, a successful obfuscation requires raising the standard deviation of the noise (e.g. surpassing the standard deviation of the signals) to generate larger noise. However, it will inevitably impact the usability of legitimate apps.

Furthermore, attackers can apply various denoise techniques to potentially raise the success rate if the original waveform is not changed by the denoise method. In our preliminary experiment, we measure the attackers' success rate by applying two denoise methods: total variation proximity operators~\cite{prox-tv} and 1D gaussian filter.
As shown in the spectrograms in Fig.~\ref{side_channel_prelim_fig}(b), the new energy components at higher frequencies are removed by denoising. In our experiment, by applying appropriate denoising methods on Mcgill and ZJU datasets, attackers can achieve higher success rates.

\textbf{Our approach.} We propose a new defense mechanism against strong adversaries by making a small extension, that supervises the neural network to learn the injected noise and use it as a hidden fingerprint. The idea is to embed a slightly noised signature in the sensing signal, and instruct the neural network to recognize such signature if the attacker launches a replay attack from the sniffed data. Specifically, our design is backed by the recent discovery that the neural networks can even fit unstructured random noise with random labels~\cite{match-noise}. It is also supported by the studies in adversarial learning~\cite{madry}, which supervises the classifier to distinguish random perturbations induced from the adversarial examples~\cite{ifgsm}. Here, our intention is that if the pair of genuine and noised data (genuine plus noise) is labeled as negative, the Siamese Network is being supervised to map them into different areas in the metric space, similar to how the adversarial examples are learned~\cite{madry}, i.e., where the ``noise'' are treated as hidden features perceivable by the neural network. Admittedly, this would make the network more difficult to train, in order to learn the patterns from the additive noise and distinguish from the genuine data. With the superb representational capability of the neural network to approximate continuous functions (the universal approximation theorem~\cite{hornik}), it is able to fit the noise part as validated in our evaluation. By labeling them as negative, the system thus forcibly learns the nuances between genuine and noised samples.

The extension can be realized by creating a simple wrapper around \texttt{SensorManager}, attackers using the noised data for replay attacks would be recognized by the system. Meanwhile, our mechanism significantly reduces the level of noise without sacrificing much usability from the benign applications, while the naive noise injection requires a large noise level for successful obfuscation. To mitigate the impact of possible denoising from the attacker, the system can further predict the potential classical denoise algorithms that might be used by attackers. The authentication module can generate the denoised pairs beforehand and similarly label them as negative for training. The detailed evaluation of our proposed defending system is presented in Section \ref{side_channel:sec}.

\begin{figure*}[!ht]
\vspace*{-0.09in}
\centering
\hspace*{-0.3in}
\begin{subfigure}[b]{0.3\textwidth}
                \includegraphics[width=1.1\textwidth]{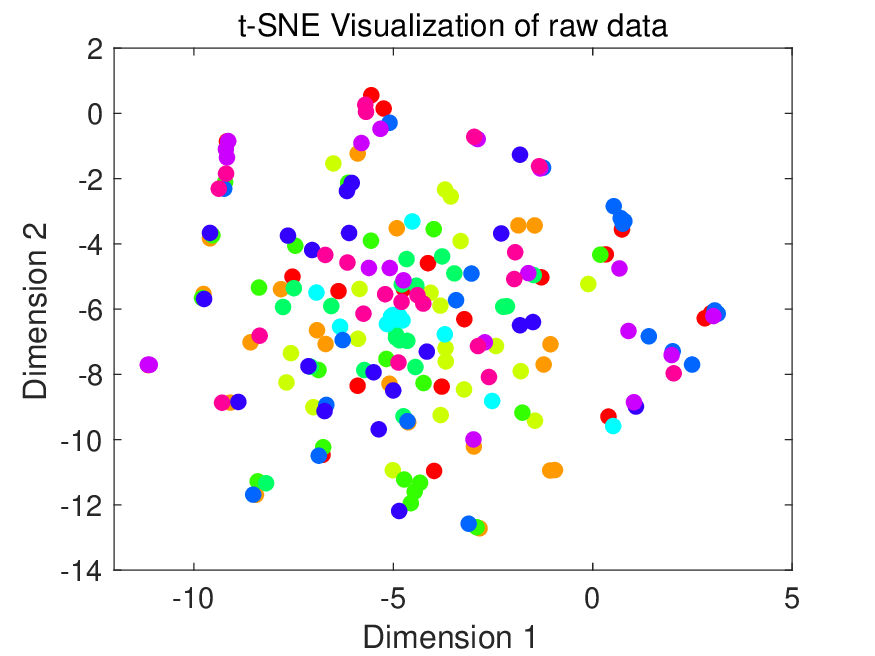}
                \vspace{-0.2in}
                \caption{}
\end{subfigure}
\hspace*{0.05in}
\begin{subfigure}[b]{0.3\textwidth}
                \includegraphics[width=1.1\textwidth]{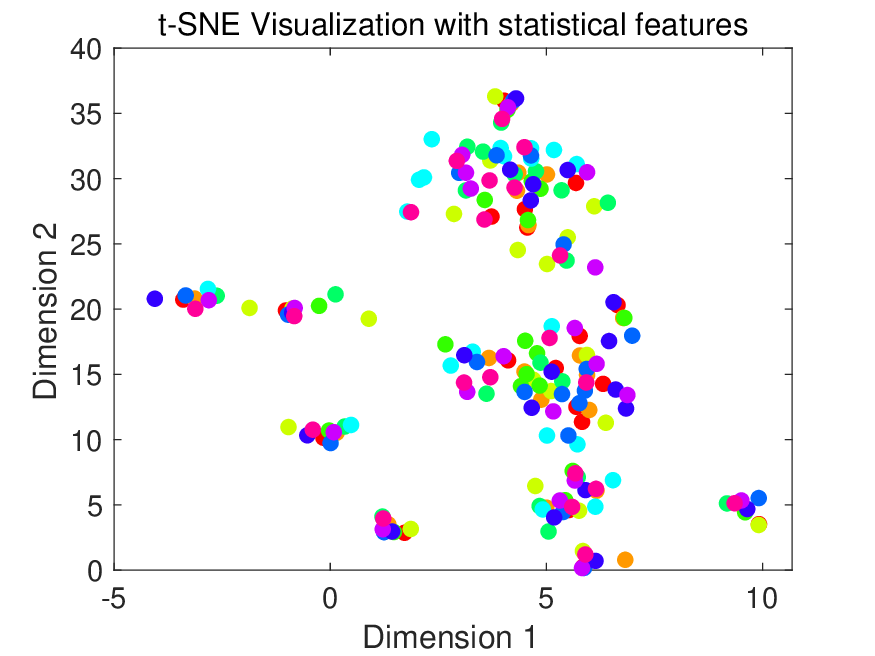}
                \vspace{-0.2in}
                \caption{}
\end{subfigure}
\hspace*{0.05in}
\begin{subfigure}[b]{0.29\textwidth}
                \includegraphics[width=1.1\textwidth]{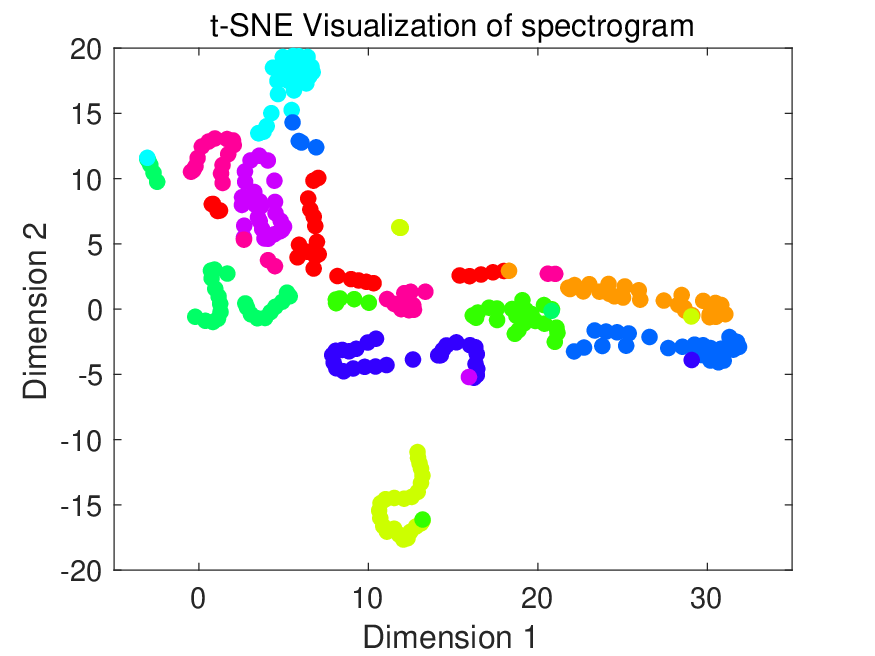}
                \vspace{-0.2in}
                \caption{}
\end{subfigure}%
\vspace*{-0.11in}
\caption{t-SNE visualization (best view in color; each color represents samples from an individual) (a) raw walking data (b) nine statistical features (c) spectrogram. }
\label{t-sne-fig}
\vspace*{-0.2in}
\end{figure*}

\vspace{-0.1in}
\subsection{Speed up Convergence via Privacy-Preserved Feature Transfer}
Training learns hundreds of thousands of parameters through backpropagation, which could take more than hundreds of epochs till convergence. The previous work proposed to partition the neural network between the cloud and mobile device in a layer-wise manner~\cite{surgeon}. We build on this approach to speed up convergence via feature transfer. With domain similarities, knowledge learned from the source can be efficiently repurposed for the target domains. For neural networks, the high-level features learned from the first few layers are more generic, while the low-level features are more specific to the classification tasks~\cite{yosinski} (e.g., the early layers learn general features like edge detectors to identify the concentration of frequency energy from the sensing signals).

To initiate, the cloud (\emph{source}) and the mobile (\emph{target}) agree on a partial network structure of the first $k$ layers, e.g., the first two convolutional layers. The agreement includes the model weights and hyper-parameters. For the target model, a few adaptation layers are introduced. Thus, the high-level features could be efficiently reused on user's mobile device. We can utilize a source model $\mathcal{M}_s$ trained on the public dataset with $n_s$ classes in the cloud, and transfer the learned features for the $n_t$ classes in the target model $\mathcal{M}_t$, when the source and target domains do not overlap.

Specifically, samples $x$ from the source domain are passed through $\mathcal{M}_s$ until the $k$-th cutoff layer, where $x$ is represented as an $n$-byte feature vector. Next, the model parameters $\mathcal{M}_s$ are transferred to the target model $\mathcal{M}_t$ for the first $k$ layers, along with all the feature vectors. To initiate training on mobile, the target model freezes weights of the first $k$ layers. The error is backpropagated from the last layer to the $(k+1)$-th layer. The weights of the adaptation layers are adjusted according to stochastic gradient descent. Note that the cloud is not aware of the layer structure or model weights beyond the $k$-th layer on the mobile devices. We show by experiments that this approach has great potentials of adaptation, such as allowing the source and target to have different loss functions (softmax for $\mathcal{M}_s$ and contrastive loss for $\mathcal{M}_t$) and heterogenous sensor hardware with different sampling frequencies. The target model can still learn effectively and enjoy massive speed-up of convergence with little accuracy loss.

\textbf{Remarks on Privacy and Communication.} Other than the active attackers, privacy exploits attempt to reconstruct the original data from feature activations~\cite{invert} or model parameters~\cite{ateniese}. Our design is robust against these exploits since: 1) activations generated from the target model are kept on mobile thus curious cloud providers cannot invert the private data; 2) though data can be recovered from the shared weights of $k$ layers on mobile by curious users, the data is public and carries little business value; 3) the model weights beyond the $k$-th layers on mobile are not disclosed to anyone else, hence a third party cannot recover private data from the model parameters.

The cost from network communication is associated with the computational speed-up. Fortunately, the neural architectures need to remain compact to fit into the device's memory, so the increased network overhead to transmit the intermediate features is manageable for on-device implementation. For T-mobile 4G LTE at 60 Mbps or the campus WiFi network at 90 Mbps downlink speed, the communication takes about 1-2 mins for the architectures in Table \ref{architecture}. Features can be downloaded in an offline fashion when the mobile device is connected to the WiFi. To further reduce communication overhead, compression can be applied to take advantage of the inherent sparsity in the feature activations, e.g., the zero-value compression~\cite{zvc} and ZLIB compression~\cite{zlib} can reduce the network bandwidth by $3-5\times$~\cite{compression}.


\vspace{-0.05in}
\section{Implementation}

\vspace{-0.05in}
\subsection{Data Pre-processing} \label{sec:spectrogram}
Data processing lays the foundation to achieve high accuracy. Here, we formally discuss the intuition behind using \emph{spectrograms} to represent sensing signals gathered by accelerometer sensors. Unlike images, the accelerometer signal is one-dimensional time series. Existing research mainly works in the time-domain and requires cycle extraction~\cite{idnet} or segmentation~\cite{deepmag}. Cycle extraction looks for cyclic patterns between local minima or maxima algorithmically, but is prone to error in the presence of noise. Segmentation divides the signal into many overlapped pieces that expands the dataset by many folds. Here, we adopt a new approach to model walking data as speech and demonstrate its advantages in the following.

Walking consists of a set of motions from the body parts (torso and limb), which shares similarities with speech from their generation mechanisms. While speaking, the pulse from vocal cords is modulated in frequency through the throat cavity and reshaped by the articulators (tongue, mouth, lips) to produce sound. Gait signals generate a similar pattern from the body parts. Based on these observations, it is reasonable to model gait as speech, which is typically analyzed in spectrogram~\cite{spectrogram}.

A spectrogram uses three dimensions to represent signal energy as a function of time (x-axis) and frequency (y-axis). It breaks data into segments of short intervals, takes short-time Fourier transform in each segment and assigns frequency spectrums into different bins of magnitude. Each bin stands for the frequency scale perceived. Spectrogram concatenates multiple quasi-stationary cycles to generate a 2D output. This way, learning can be performed effectively using convolutional neural networks.

The compelling advantage of spectrogram is suggested by Fig.~\ref{t-sne-fig} (visualized by the t-SNE tool to reduce the high-dimensional data into 2D~\cite{t-sne}). Fig.~\ref{t-sne-fig}(a) shows the raw sensing signal and Fig.~\ref{t-sne-fig}(b) visualizes the data with nine statistical features. Though these statistical features form a distinguishable trend, they are still not powerful enough; in sharp contrast, data points are clustered in a more organized manner in Fig.\ref{t-sne-fig}(c), so it is much easier to build a classifier and recognize different individuals.

\vspace{-0.05in}
\subsection{Model Development}

Model architecture determines the representational power, memory requirement, and computational intensity. In this paper, we evaluate three convolutional neural network architectures extended the families of \textit{LeNet}~\cite{lenet}, \textit{VGG}~\cite{vgg} and \textit{MobileNetv2}~\cite{mobilenetv2}. Though an alternative is to use the recurrent neural networks~\cite{motion}, the computation intensity is much higher. We customize these classic models to add or prune layers in order to yield similar input dimension at the dense layer as their original implementation with the ImageNet. The spectrograms of $(x,y,z)$ axis are stacked vertically to form $33\times42$ images.

In particular, LeNet repeats two blocks of $5\times5$ convolutional and max pooling layers followed by densely connected layers. We add one more $3\times3$ convolutional layer and prune one dense layer to get $4$ weight layers (therefore the name LeNet4). VGG repeats two $3\times3$ convolutional layers to achieve similar receptive field with the $5\times5$ convolution, but much less computation/parameters. Multiple such blocks are stacked to learn complex relations among the features. We repeat the blocks three times and introduce one more convolutional layer before the last max pooling, thus making VGG8 a heavy-weight network with 8 weight layers. We also implement the latest MobileNetv2. The model stacks inverted residual blocks (inv\_res\_bl) to take low-dimensional representation, and then expands to the high-dimension for efficient feature extraction by the depthwise convolution. The blocks are connected with bypass links to make deeper structures trainable with less degradations. The max pooling layer is replaced by a convolution stride of $2$, e.g. $(1,1,2)$ represents two blocks with a stride of $1$ followed by a block with a stride of $2$. Table~\ref{architecture} summarizes the model architectures and layer-wise parameters.

\vspace{-0.05in}
\subsection{Mobile Development}
The choice of the software framework is crucial since training requires backpropagation. Despite a handful of available frameworks, most of them (e.g. \emph{Tensorflow Lite}~\cite{tensorflow,caffe2,mxnet}) have tailored backpropagation and left only the inference part to compute from pre-trained models. This way, no intermediate gradient values need to be stored and the memory/code can be optimized. In this paper, to enable training on the mobile device, we develop the system on a Java-based framework called \emph{DL4J}~\cite{dl4j}. Since the two Siamese branches are identical, only one copy of the model is stored in memory. During testing, we notice that deeper structures could cause \texttt{OutOfMemoryError} due to a large number of parameters and batched data processing. To mitigate, we set \texttt{largeHeap} to give the application a 512 MB heap capacity.

\begin{table*}[t]
\vspace{-0.08in}
\small
\begin{tabular}{|c|c|c|c|c|c|}
\hline
\multicolumn{2}{|c|}{LeNet4} &
\multicolumn{2}{|c|}{VGG8} &
\multicolumn{2}{|c|}{MobileNetv2} \\
\hline
\#layer blocks & \# param.     &\# layer blocks & \# param.                  &\# layer blocks & \# param. \\
\hline
$32\times$conv2d$(5,5)+$pool & 0.96K     & $2\times(64\times$conv2d$(3,3)+$pool$)$  &39.2K       & conv2d$(3,3,2)$ & 1.9K \\
$64\times$conv2d$(5,5)+$pool & 51.5K     & $2\times(128\times$conv2d$(3,3)+$pool$)$ &222.5K      & $(16,32)\times$inv\_res\_bl$(1,1,2)$ & 47.3K \\
$32\times$conv2d$(3,3)$ & 51.4K             & $3\times(128\times$conv2d$(3,3)+$maxpool$)$ &444.3K      & conv2d$(1,1)$ & 1.1K \\
dense$(128)$ & 82.5K             & dense$(128)$ & 327.8K           & dense$(64)$ & 61.8K \\
\hline
contrastive/x-entropy loss & 186.36 K             & contrastive/x-entropy loss  &1033.8K           & contrastive/x-entropy loss & 112.1 K \\
\hline
\end{tabular}
\vspace{-0.1in}
\caption{Summary of model architectures} \label{architecture}
\vspace{-0.1in}
\end{table*}

\vspace{-0.05in}
\section{Experimental Evaluation}  \label{evaluation:sec}
This section conducts a thorough evaluation of the framework for deep behavioral authentication. The main goals of the evaluations are: 1) investigate the accuracy and computational cost of different models and approaches; 2) examine cost savings and performance impact from feature transfer; 3) validate system robustness against both random and active attacks; 4) profile performance and overhead on various smartphone models.

\begin{figure*}[!t]
\vspace*{-0.09in}
\centering
\hspace*{-0.3in}
\begin{subfigure}[b]{0.25\textwidth}
                \includegraphics[width=1.1\textwidth]{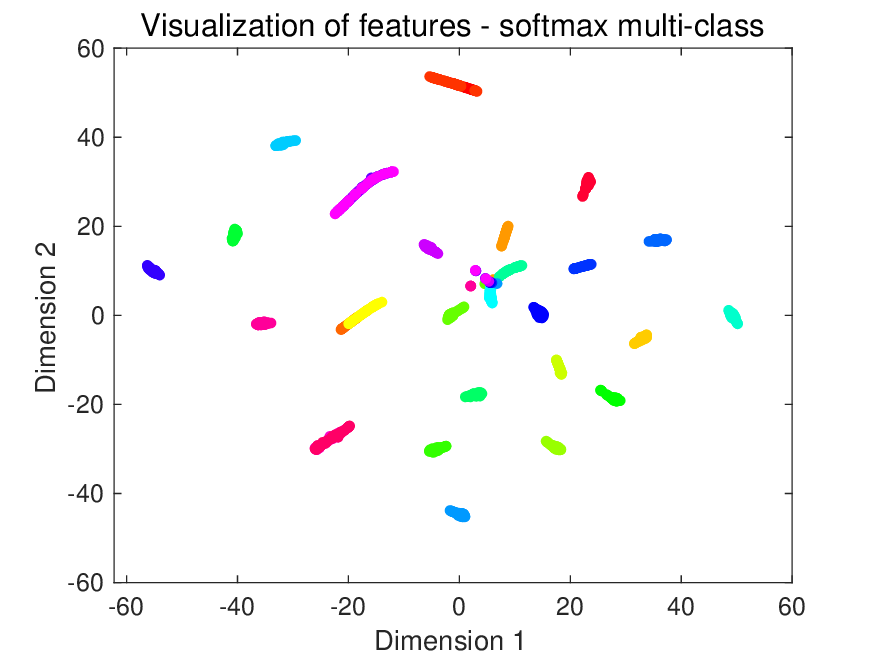}
                \vspace{-0.2in}
                \caption{}
\end{subfigure}
\begin{subfigure}[b]{0.25\textwidth}
                \includegraphics[width=1.1\textwidth]{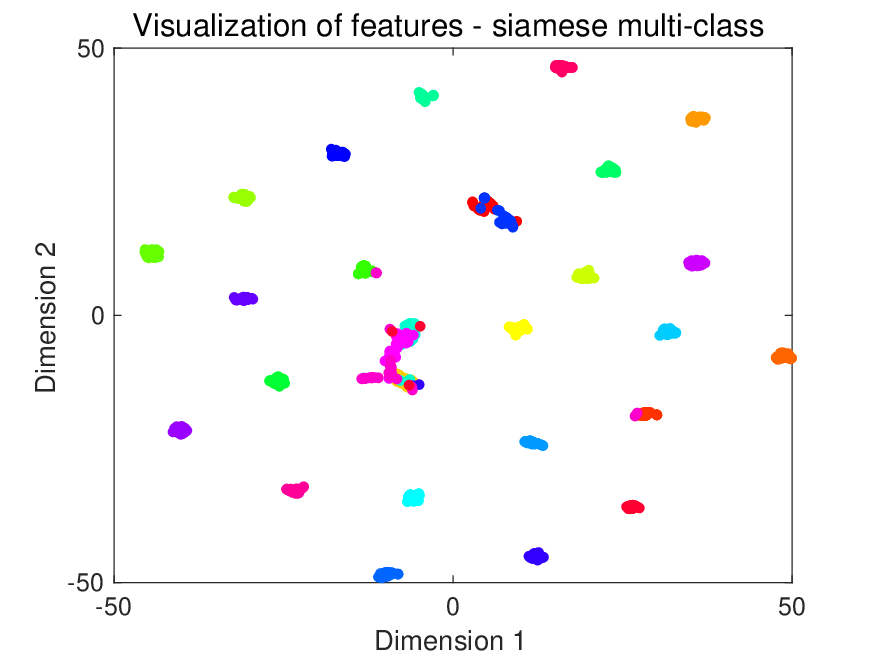}
                \vspace{-0.2in}
                \caption{}
\end{subfigure}
\begin{subfigure}[b]{0.25\textwidth}
                \includegraphics[width=1.1\textwidth]{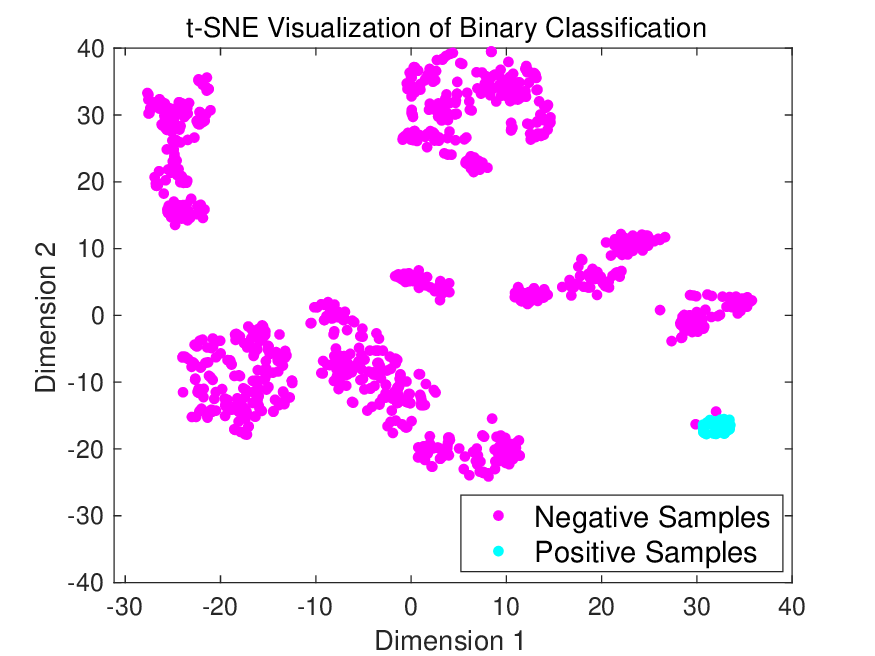}
                \vspace{-0.2in}
                \caption{}
\end{subfigure}
\begin{subfigure}[b]{0.25\textwidth}
                \includegraphics[width=1.1\textwidth]{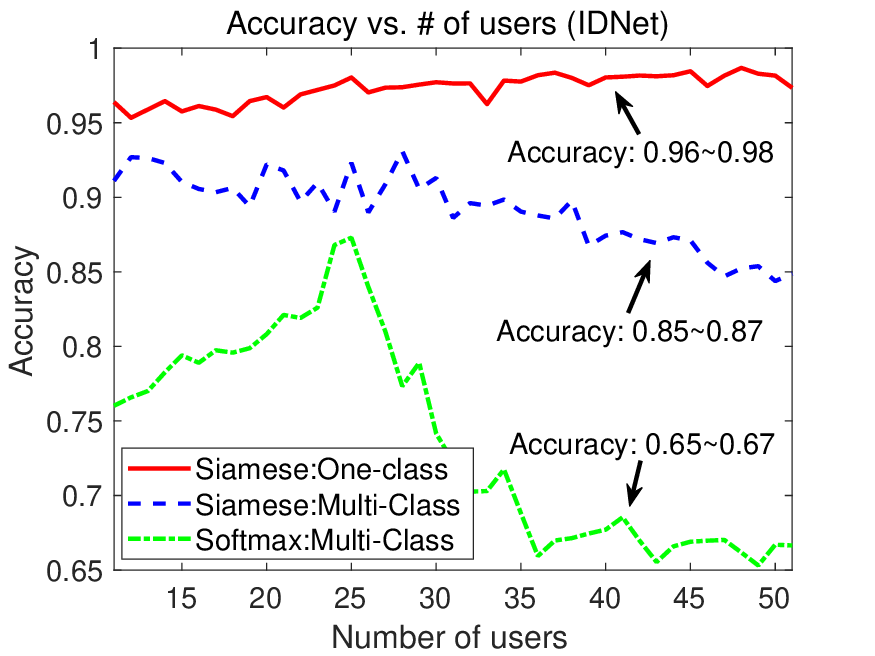}
                \vspace{-0.2in}
                \caption{}
\end{subfigure}%
\hspace*{-0.2in}
\vspace*{-0.11in}
    \caption{Multi-class and binary classification via t-SNE visualizations (a) softmax (multi-class); (b) siamese (multi-class); (c) siamese (binary); (d) accuracy vs. user number. (Best view in color)}
\label{feature_fig}
\vspace*{-0.07in}
\end{figure*}

\vspace{-0.05in}
\subsection{Dataset and Experimental Settings}
To make the benchmarks comparable, the experiments are based on public datasets: Mcgill~\cite{mcgill_dataset}, IDNet~\cite{IDnet_dataset}, ZJU~\cite{zju_dataset} and Osaka~\cite{osaka_dataset} gait datasets. Note that this paper focuses on algorithm design and system integration rather than collecting, analyzing or deriving data from human subjects. Thus, an IRB approval is not required. With a total coverage of around 1,000 individuals, we believe the four datasets are sufficient to validate the system in various scenarios.

In particular, Mcgill includes $15$-min walk of $20$ people on two different days. IDNet is collected in a more vibrant environment with 50 participants, where there is no restriction of phone types and clothes (different clothes would lead to slightly different gait patterns from the same individual). ZJU collects gait data from 153 individuals in 3 different sessions using 5 body sensors of low sampling rates. Osaka records $1$-minute walk of $744$ subjects. Due to short recordings (only $1$-$2$ spectrograms), we cannot perform meaningful training so it is utilized as a large database from which attackers may launch random attacks. Since some individuals have much less or missing data in IDNet and ZJU, we remove those individuals for data balance. This ultimately brings them to $30$ and $136$ individuals respectively.

\begin{table*}[!t]
\vspace{-0.08in}
\centering
\small
\begin{tabular}{@{}l l c c c c c c c c c@{}}
\toprule
&\multirow{2}[3]{*}{} & \multicolumn{3}{c}{baseline} & \multicolumn{3}{c}{siamese multi-class} & \multicolumn{3}{c}{siamese binary-class (20\% data)} \\
\cmidrule(lr){3-5} \cmidrule(lr){6-8} \cmidrule(lr){9-11}
& &softmax(sw) &softmax(spgm) &osvm & contrastive & x-entropy & joint & contrastive & x-entropy &joint\\
\midrule
\multirow{4}{4mm}{\begin{sideways}\parbox{7mm}{Mcgill}\end{sideways}}
&LeNet4 &0.774 &0.881 &0.542 & 0.918 & 0.940 &0.925 &0.966 &0.934 &\textbf{0.975}\\
&VGG8 &0.752 &0.902 &0.672 & 0.925 &0.952 &0.931 & 0.962 &0.906 &\textbf{0.973} \\
&Mobilenet &0.682 &0.811 &0.581 & 0.865 & 0.926 &0.923 &0.847 &0.901  &\textbf{0.957} \\ \hline
\multirow{4}{4mm}{\begin{sideways}\parbox{6mm}{IDNet}\end{sideways}}
&LeNet4 &0.726 &0.842 &0.552 & 0.884  & 0.903 &0.910 &0.937 &0.899 &\textbf{0.943} \\
&VGG8 &0.764 &0.875 &0.561   & 0.916 & 0.934 &0.915 &0.908  &0.901 &\textbf{0.941} \\
&Mobilenetv2 &0.770 &0.776 &0.591  & 0.876 & 0.912 &0.867 &0.910 &0.921 &\textbf{0.945}\\ \hline
\multirow{4}{3mm}{\begin{sideways}\parbox{5mm}{ZJU}\end{sideways}}
&LeNet4 &0.442 &0.646 &0.511  & 0.681  & 0.804 &0.779 &0.941 &0.926 &\textbf{0.972}\\
&VGG8 &0.463 &0.743 &0.523 & 0.769  &0.841 &0.800 &0.936 & 0.851 &\textbf{0.981}\\
&Mobilenetv2 &0.591 &0.471 &0.510 & 0.706  &0.778 &0.743 &0.895 & 0.835 &\textbf{0.921}\\
\bottomrule
\end{tabular}
\vspace{-0.1in}
\caption{Model accuracy of different loss functions for the siamese network}  \label{accuracy}
\vspace{-0.1in}
\end{table*}

The datasets are first split into 80\% for training and 20\% for testing. Then Algorithm \ref{sampling_algo} is adopted to form balanced positive and negative pairs from the training and testing set respectively. For SPRT test, the pairs are generated by randomly pairing training samples with testing samples as described in Section \ref{sec:sprt}, where the training samples act as the ground truth. This simulates the run-time when new motion data is evaluated against training samples as the ground truths. To assess the performance of authentication, we mainly focus on the mean Average Precision (mAP), which is the average percentage of true authentication over the total number of testing. We also evaluate the trade-offs between false rejection (the genuine user is falsely rejected) and false acceptance (an imposter is falsely accepted) using different margin threshold.

We set the margin $m=1.5$ in the contrastive loss (Eq.~(\ref{eq_contrastive})) based on the best Equal Error Rate from the experiment discussed in Section \ref{sec:miscellaneous}. $\gamma=0.1$ is set empirically for the joint loss (Eq.~(\ref{eq_joint_loss})) so the training is led by the contrastive loss. For fast prototyping, we first develop the model and evaluate authentication accuracy, security and performance in \emph{Tensorflow}~\cite{tensorflow} with Nvidia Tesla P100 GPU, and then develop the learning module on Nexus 6/6P, Huawei Mate 10 and Google Pixel2 using DL4J~\cite{dl4j}. A large batch size of $128$ is used while training on GPU for 100 epoches. The learning rate is set to $0.03$ with the RMSprop optimizer. Accuracy is averaged over 10 different runs and each run draws a random subset of samples to construct balanced paring between positive and negative pairs following Algorithm \ref{sampling_algo}. During our testing, we find that the maximum batch size for Nexus 6 (oldest phone in our test) is $56$ pairs. To test various models and avoid memory errors, we set the batch size to $20$ on mobile.

\subsection{Authentication Accuracy}   \label{sec:acc}
We first evaluate the authentication accuracy by comparing models, data representation and learning mechanisms on different datasets in Table~\ref{accuracy}. We validate the choice of spectrogram by comparing with the pre-processing technique of sliding window (SW)~\cite{deepmag} on the temporal data both using softmax (cols. 1, 2). As envisioned by the t-SNE visualization of Fig.\ref{feature_fig}, spectrogram achieves a significant accuracy gain of over 10\% (col. 4). A one-class SVM (osvm) is used in~\cite{idnet} to detect outliers from imposters. It takes features from the last convolutional layer learned from the softmax function to train an osvm using positive samples only. Unfortunately, though osvm can handle 80-90\% outliers, it fails to generalize to the positive samples, which results in high rate of false rejections. Thus, the total accuracy is just slightly better than random guesses (col. 3).

\textbf{Softmax vs. Contrastive Loss.} Our motivation to use the Siamese Network is because of the higher discriminative power on small data. To validate, we first visualize the features learned by softmax and siamese (contrastive loss) in Figs.~\ref{feature_fig}(a) and (b), where the colors represent the feature vectors of different subjects in 2D. Features learned by softmax are not sufficiently discriminative where the distance along the feature vectors from the same individual could be similar to a different individual. We further notice that some features belong to different individuals are mapped to the same vector space in 2D. These findings are in line with~\cite{centerloss} (softmax tends to underperform). Contrastive loss from the siamese network offers improvements by mapping feature activations into a condensed, compact set of spaces. This validates the higher discriminative power of deep metric learning than softmax especially with less training data. Not only via feature visualization, the authentication accuracy also indicates 8-15\% improvements between the two methods (col. 2 and 4-6 of Table~\ref{accuracy}).


\textbf{Binary vs. Multi-class Classification.} We discuss the impact of formulating the problem into either the binary or multi-class classification problem. Multi-class classification requires all the pairs between different classes to be labeled whereas binary only labels one vs. the rest. The former is more suitable for recognition tasks where a centralized model is trained to identify different users. The recognition model can be also migrated to the mobile devices for authentication~\cite{idnet}. However, it is subject to potential security risks when a malicious end user attempts to invert training data of other individuals ~\cite{ateniese}. In addition, we investigate the performance gap between the two formulations in terms of model accuracy.

To simulate limited mobile storage, only 20\% data from the training set is used for binary classification but evaluated on the entire test set. This is challenging for recognition since the neural network can only ``see'' from a small subset of training data. A model is trained for each individual and the results are averaged. Fig.~\ref{feature_fig}(c) visualizes binary classification. It only distinguishes the positive samples from the rest and the negative samples can be mapped to similar locations in space without causing an error. Nevertheless, multi-class classification still has to separate all the individuals by a margin, which makes it difficult to differentiate hard samples (Figs.~\ref{feature_fig}(a)(b)). To test the scalability of multi-class classification, we show the results in Fig.~\ref{feature_fig}(d) by increasing the number of classes in IDNet. The accuracy declines with a growing number of classes in the system. Hence, model capacity should keep growing as new users subscribe to the service. This would require extensive maintenance efforts in distributed mobile environments. As projected in Fig.~\ref{feature_fig}(d), accuracy is independent from the system scale using binary classification with a fixed network architecture.

Table 2 summarizes the overall accuracy comparison. With multi-class classification using the Siamese Network, accuracy still declines a little with an increasing number of classes (e.g. from $0.952$ of Mcgill with 20 people down to $0.841$ of ZJU with 136 people). By reducing the problem into binary classification, the accuracy stays above 90\%. Among them, the new joint loss accomplishes the best accuracy with over 95\% correctness. This is because the joint loss balances the two loss functions and combines the model outputs for higher fidelity.

We also notice some interesting phenomenon that the cross-entropy loss is better than the contrastive loss for multi-class classification, but the opposite for binary classification. The difference between them is that the cross-entropy generates a probabilistic decision, rather than a deterministic distance metric from the contrastive loss. In our experiment, we discover that contrastive loss is more prone to error during multi-class classification in the presence of hard samples. Due to space limit, we would further investigate this issue in our future work. Finally, we alter the model into VGG8 and MobileNetv2. VGG8 achieves the best accuracy in most cases. With 40\% less parameters, MobileNetv2 suffers 8-26\% accuracy loss compared to LeNet4. This indicates that networks particularly optimized on model parameters and computer vision tasks may perform poorly on mobile sensing tasks, compared to simple solutions of stacking convolutional layers such as VGG8.

\subsection{Resource Requirement}

\begin{figure*}[!t]
\vspace*{-0.09in}
\centering
\hspace*{-0.2in}
\begin{subfigure}[b]{0.25\textwidth}
                \includegraphics[width=1.05\textwidth]{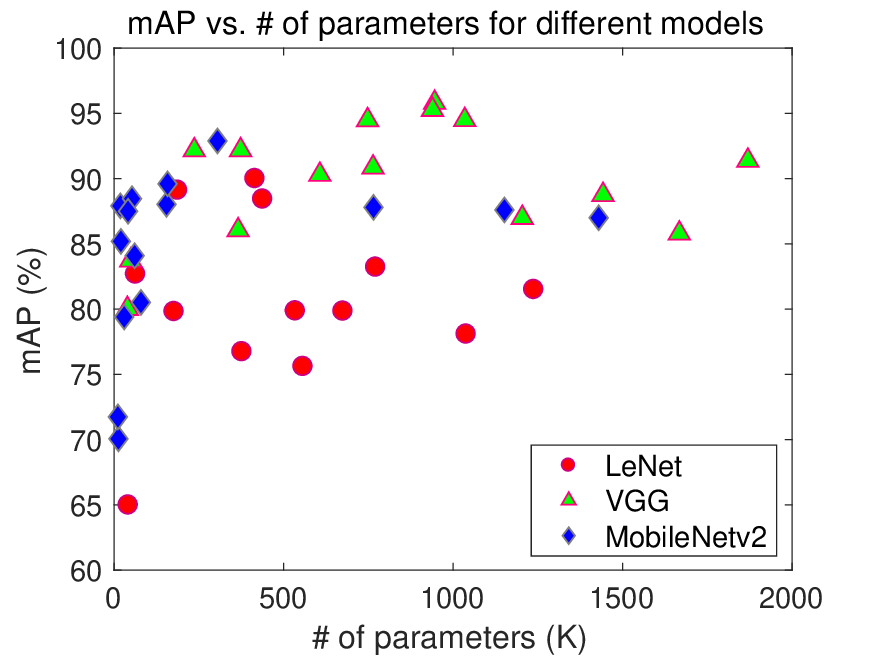}
                \vspace{-0.21in}
                \caption{}
\end{subfigure}
\begin{subfigure}[b]{0.25\textwidth}
                \includegraphics[width=1.05\textwidth]{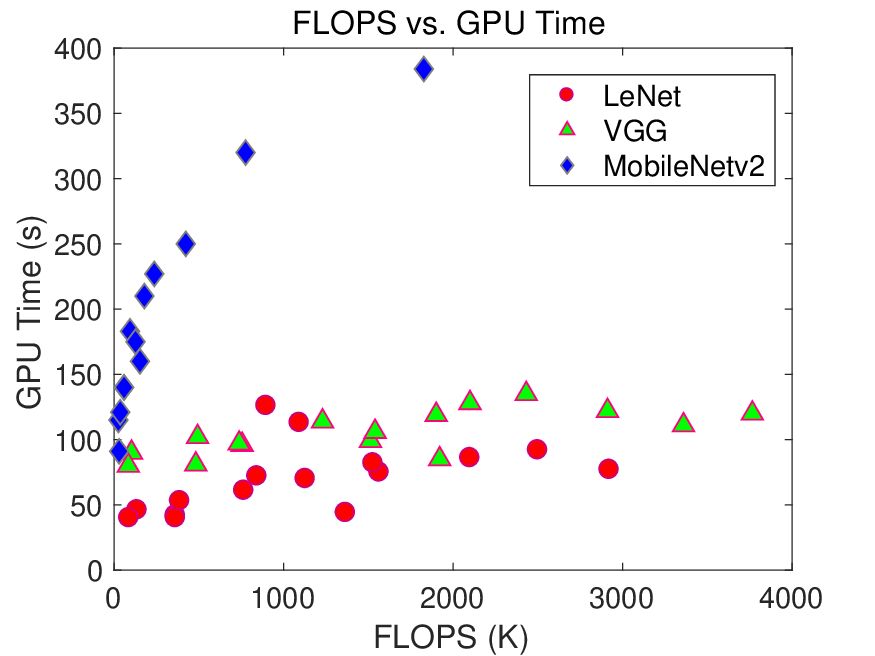}
                \vspace{-0.21in}
                \caption{}
\end{subfigure}
\begin{subfigure}[b]{0.25\textwidth}
                \includegraphics[width=1.05\textwidth]{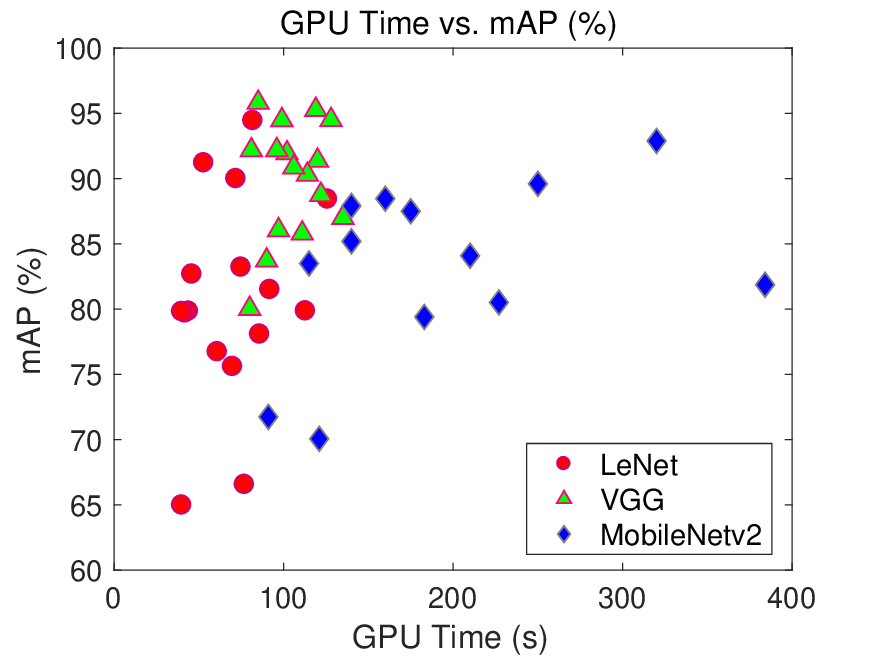}
                \vspace{-0.21in}
                \caption{}
\end{subfigure}
\begin{subfigure}[b]{0.25\textwidth}
                \includegraphics[width=1.05\textwidth]{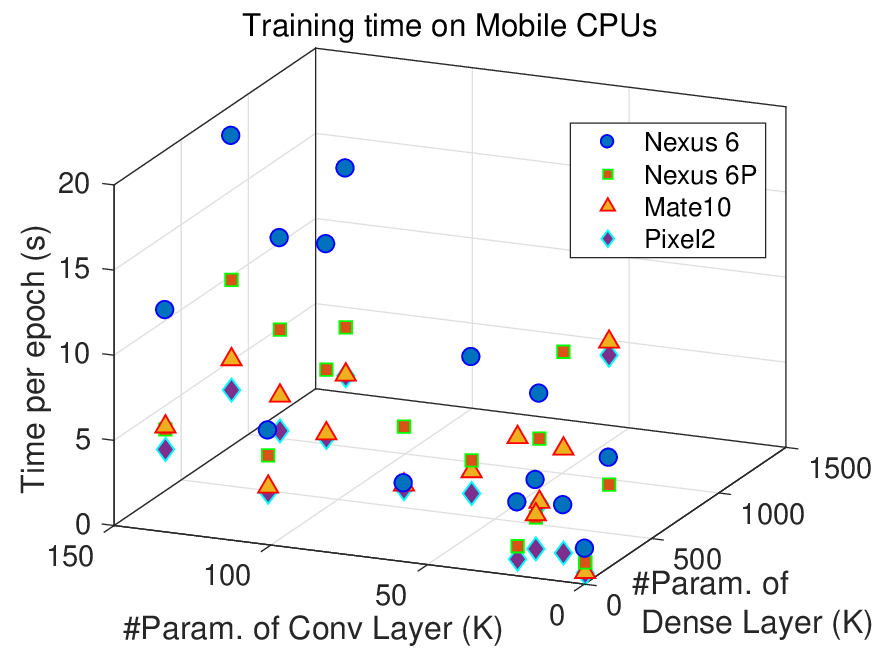}
                \vspace{-0.21in}
                \caption{}
\end{subfigure}%
\hspace*{-0.01in}
\vspace*{-0.13in}
\caption{Evaluation of resource requirement vs. accuracy on GPU and mobile platforms using IDNet (a) mAP vs. parameters; (b) FLOPS vs. GPU time; (c) GPU time vs. mAP; (d) Parameters (Conv and Dense Layers) vs. Mobile CPU Time.}
\label{parameter_fig}
\vspace*{-0.15in}
\end{figure*}


To quantify the performance and resource requirements of the mobile sensing task, we conduct more experiments to illustrate the relations between model parameters, floating point operations (FLOPS), and accuracy in Fig.~\ref{parameter_fig}. We alter the structures by shrinking/expanding filter size, numbers, and adding/removing convolutional or pooling layers. For the same model, in general, more parameters bring higher representational power at the risk of overfitting and cost of computation. From Fig.~\ref{parameter_fig}(a), VGG8 is more stable than others in terms of accuracy. Once the number of parameters exceeds a million, the models tend to overfit. Mobilenetv2 can be tailored to only weigh half of LeNet4, but the performance is not stable. Fig.~\ref{parameter_fig}(b) also indicates that it incurs nontrivial GPU time if the FLOPS increase. Fig.~\ref{parameter_fig}(c) shows that LeNet4/VGG8 are more competitive than Mobilenetv2 for the datasets in terms of computation time and accuracy.

To facilitate mobile development, we conduct the following experiments using LeNet4 and keep the consistency through the rest of the experiments. Fig.~\ref{parameter_fig}(d) shows the training time per epoch on mobile devices. We plot in 3D for better visualization of the impact from the convolutional and dense layer. Training on mobile devices is not only feasible, but actually much faster than expected. For a deep model with 650K parameters and $400$ samples, it only takes the latest Pixel2 or Mate10 less than $5$ seconds to complete one training epoch. Thus, training $100$ epochs takes less than $10$ mins. Even the old Nexus 6 finishes around $10$ seconds per epoch. During the experiment, we notice that the speed bottleneck of convolutional layers is magnified on mobile devices due to less processing power from the mobile CPUs and memory. As observed in Fig.~\ref{parameter_fig}(d), with more convolutional layers, training time surges sharply. However, increasing computations of the dense layer has less impact on performance. Interestingly, we are even able to train some networks with over a million parameters, as long as most of the parameters reside in the dense layer. Equipped with the capability to learn, model updates can be scheduled efficiently without external efforts from service providers.

\subsection{Speed up on Mobile by Feature Transfer}

\begin{table}[b]
\vspace{-0.08in}
\centering
\small
\begin{tabular}{|l|l||c|c|c|c|}
\hline
\multicolumn{2}{|c|}{feature transfer} & Mcgill & IDNet & ZJU & gain/loss \\\hline
\multirow{4}{4mm}{\begin{sideways}\parbox{9mm}{Mcgill}\end{sideways}}
&\emph{fconv1-3} & 0.933 & 0.903 &0.907   &-5.2\% \\
&\emph{fconv1-2} & 0.948 & 0.927 &0.918  &-3.5\%   \\
&\emph{fconv1} & 0.953 & 0.941 &0.948   & -1.9\% \\
&gain/loss & -2.1\%  & -4.2\%  &-4.2\%  &-- \\ \hline
\multirow{4}{4mm}{\begin{sideways}\parbox{9mm}{IDNet}\end{sideways}}
&\emph{fconv1-3} &0.876  &0.941  &0.896 & -3.3\%   \\
&\emph{fconv1-2} &0.922  &0.951  & 0.911 &-0.9\%  \\
&\emph{fconv1} &0.933  &0.957  &0.936 &+0.5\%  \\
&gain/loss & -2.7\%   &+1.3\%   &-2.3\%  &-- \\
\hline
\multirow{4}{4mm}{\begin{sideways}\parbox{9mm}{ZJU}\end{sideways}}
&\emph{fconv1-3} &0.808  &0.810  & 0.829 & -12.5\% \\
&\emph{fconv1-2} &0.836  &0.818  &0.833  &-11.3\% \\
&\emph{fconv1} &0.832  &0.804  &0.847 &-11.3\%  \\
&gain/loss &-11.6\%  &-13.0\%   &-10.5\%  &-- \\ \hline
\end{tabular}
\vspace{-0.1in}
\caption{Accuracy with feature transfer}
\label{transfer_table}
\vspace{-0.1in}
\end{table}


\begin{figure}[b]
\vspace*{+0.03in}
\centering
\hspace*{-0.24in}
\begin{subfigure}[b]{0.23\textwidth}
                \includegraphics[width=1.18\textwidth]{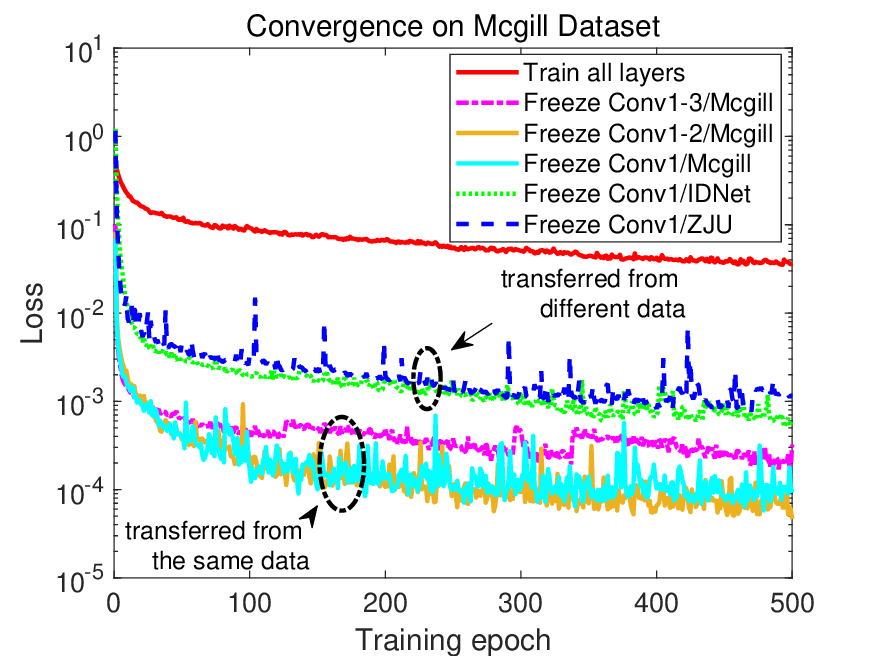}
\end{subfigure}
\hspace*{0.1in}
\begin{subfigure}[b]{0.23\textwidth}
                \includegraphics[width=1.18\textwidth]{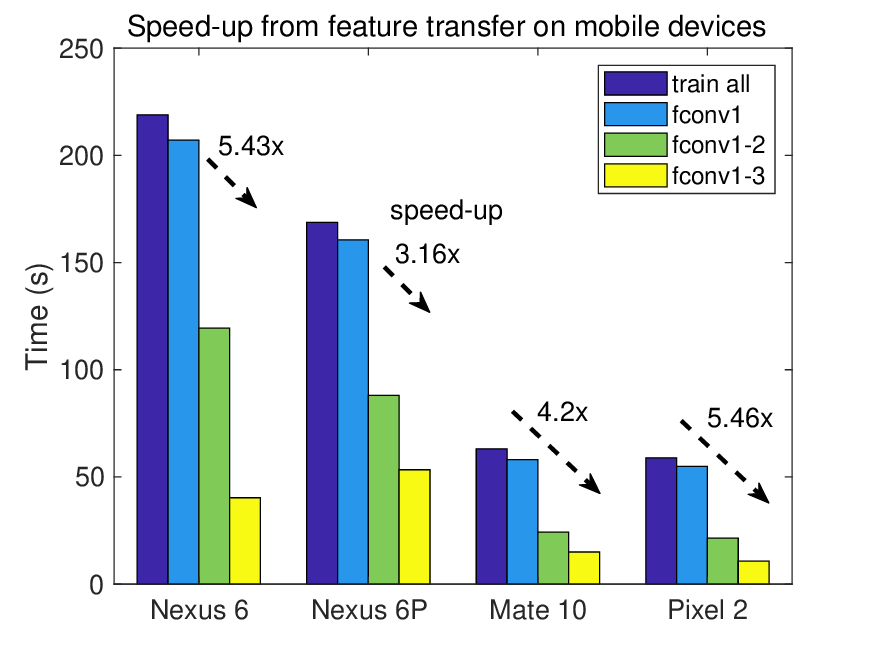}
\end{subfigure}%
\vspace*{-0.06in}
    {\small \\\hspace{0.01in} \textbf{(a) \hspace{1.55in}(b)}}
\vspace*{-0.07in}
\caption{Boost from feature transfer (a) speed of convergence; (b) speed-up on mobile devices.}
\label{transfer_fig}
\vspace*{-0.04in}
\end{figure}

Since convolutional layers learn common features, these features can be efficiently transferred from the cloud for computation efficiency. To see such potential, the following cases are evaluated: 1) freeze all convolutional layer weights (\emph{fconv1-3}); 2) freeze first two convolutional layer weights (\emph{fconv1-2}); 3) freeze the first convolutional layer weights (\emph{fconv1}). We train the rest of the layers. The source model conducts multi-class classification on the dataset (public) without the presence of the target user (private). At the target user, it performs the binary classification based on the weights transferred from the source model. Note that this implementation is robust against privacy exploits since the private activations are kept on mobile and the transferred features are public. We also evaluate scenarios when different public data are available, by alternating the source data between the other two datasets. This allows us to examine the generality of features and their impact on accuracy and convergence. If the source and target models permits easy domain adaptations, the cloud no longer needs to tightly match the hardware configuration with the user device.

Fig.~\ref{transfer_fig}(a) shows the convergence of a random individual from the Mcgill dataset. We can see that feature transfer offers at least two orders of magnitude speed-up in terms of convergence. Features learned from data gathered with different settings offer significant boost as well. For instance, for the loss value to converge to $0.05$, the original training takes $325$ epochs. With feature transfer, it only takes $2$ epochs from the same dataset, $5$ and $4$ epochs for different IDNet and ZJU datasets, respectively.
We then evaluate the speed-up on mobile devices and measure the total computation time to finish $50$ epochs of training, as shown in Fig.~\ref{transfer_fig}(b). Freezing all the convolutional layers offers $3$-$5$ times of speed-up. If one additional convolutional layer is released, the gain is still over $2$ times. The speed-up comes with a little accuracy loss due to the discrepancy among domain features (illustrated in Table~\ref{transfer_table}). Training the dense layers only has $3$-$5$\% accuracy loss on Mcgill, IDNet, and 12\% on ZJU dataset. The accuracy can be improved by fine-tuning more layers (e.g. to 0.9\% and 3.5\% for Mcgill and IDNet). Transferring from a different dataset only incurs minor accuracy loss ($1$-$3$\% on average). This indicates that the proposed architecture is robust to re-use features for the new target domain, though device settings such as sampling frequency (sensors) can be different.


\subsection{Robustness against Intra-class Variations}
\begin{figure}[t]
\centering
\hspace*{-0.3in}
\begin{subfigure}[b]{0.23\textwidth}
                \includegraphics[width=1.15\textwidth]{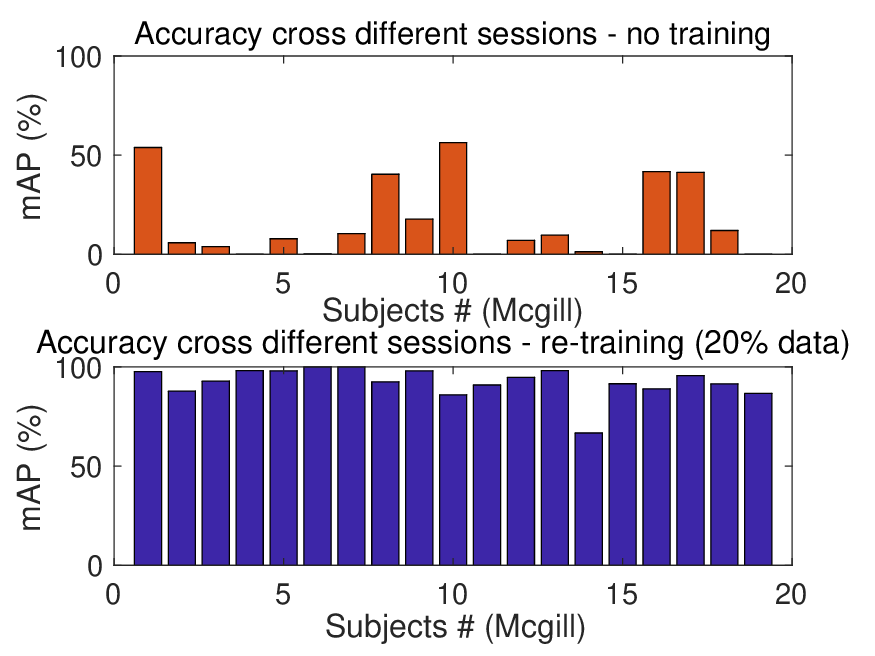}
\end{subfigure}
\hspace*{0.1in}
\begin{subfigure}[b]{0.23\textwidth}
                \includegraphics[width=1.15\textwidth]{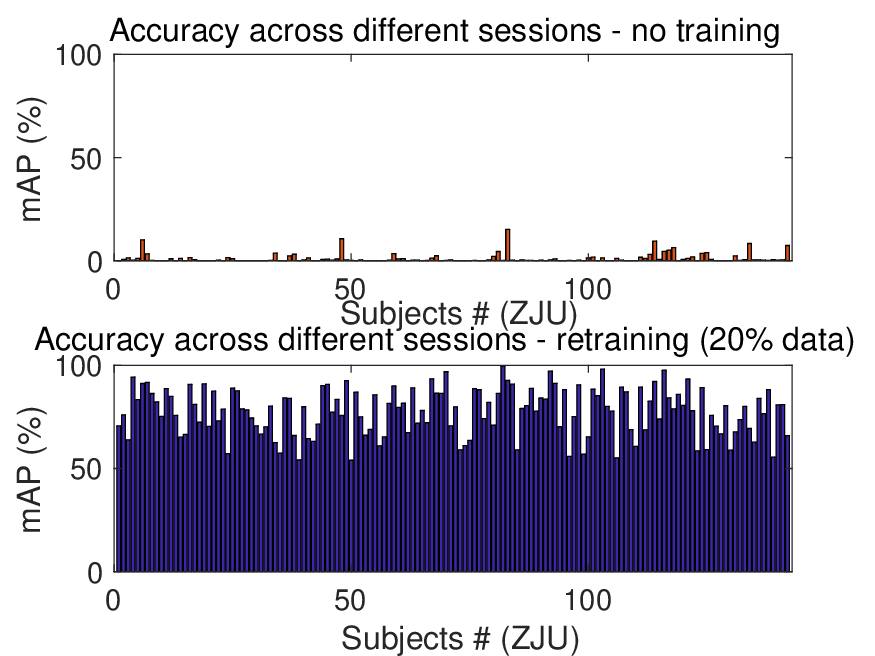}
\end{subfigure}%
\vspace*{-0.05in}
    {\small \\\hspace{0.01in} \textbf{(a) \hspace{1.55in}(b)}}
\vspace*{-0.1in}
\caption{Acceptance rate across different sessions (a) Mcgill; (b) ZJU.}
\label{intersession_fig}
\vspace*{-0.05in}
\end{figure}

We show that incorporation of training on mobile devices offers fast response to intra-class variation when behavioral biometrics evolve. We utilize the Mcgill and ZJU datasets since they record more than two sessions of a subject on different days (Mcgill) and months (ZJU). To see whether the system can still recognize its owner, we examine the acceptance rate. If the acceptance rate is low, the model is likely to reject the genuine user and degrade usability significantly. In the upper figures (\emph{no training}) of Fig.~\ref{intersession_fig}, each user trains a model in session $1$ and directly tests on the data from session $2$. As we observe, the acceptance rate is quite low if the model is not updated. Mcgill dataset across several days only yields 16.3\% average acceptance, and the rate drops to 1.1\% for ZJU over a longer period. It certainly indicates that pre-trained models cannot adapt to new data distributions.

With continuous model updates, we fine-tune the model from the previous weights with a lower learning rate, and only use 20\% of the new data. The bottom figures in Fig.~\ref{intersession_fig} shows the mean acceptance percentage over all fine-tuning epochs, which quickly brings it back to 92.4\% and 77.6\% for Mcgill and ZJU, respectively. The best acceptance percentage of some users can hit 100\% indicating that the fine-tuned model can almost perfectly adapt to the new data.


\vspace{-0.03in}
\subsection{Robustness against Random Attacks}
A \emph{random attacker} tries to gain system access using his own walking data (gait) or data retrieved from a large database. Since behavioral patterns are extremely difficult to mimic by observation, we use Osaka as the database to launch attacks. These samples are entirely new to the model from unknown data distributions.
We train users in the three datasets and enumerate through all the attacking samples (1684 spectrograms) for each user. As shown in Table~\ref{random_attack}, the success ratio is below 3\%. Once the results are fused with $32$ samples randomly selected from the training data, the ratio further declines to 1\% in the worst case. This rate could be easily reduced to zero by incorporating high-level security mechanisms such as limiting the number of trials.

\begin{table}[!t]
\centering
\small
\begin{tabular}{|c|c|c|c|c|c|}
\hline
Dataset & All &Batch 4 &Batch 8 &Batch 16 &Batch 32 \\ \hline
Mcgill &0.05\%  &0.003\%  &0.003\%  &0.000\% &0.000\%   \\
IDNet &2.36\% &2.18\%  &2.014\%  &1.682\%  &1.024\% \\
ZJU &0.346\%  &0.028\%  &0.010\%  &0.004\%  & 0.001\% \\  \hline
\end{tabular}
\caption{Success ratio of passive attacks using Osaka dataset} \label{random_attack}
\vspace{-0.1in}
\end{table}

\vspace{-0.1in}
\subsection{Robustness against Active Attacks}  \label{side_channel:sec}
\begin{figure*}[!ht]
\centering
\hspace*{-0.3in}
\begin{subfigure}[b]{0.28\textwidth}
                \includegraphics[width=1.05\textwidth]{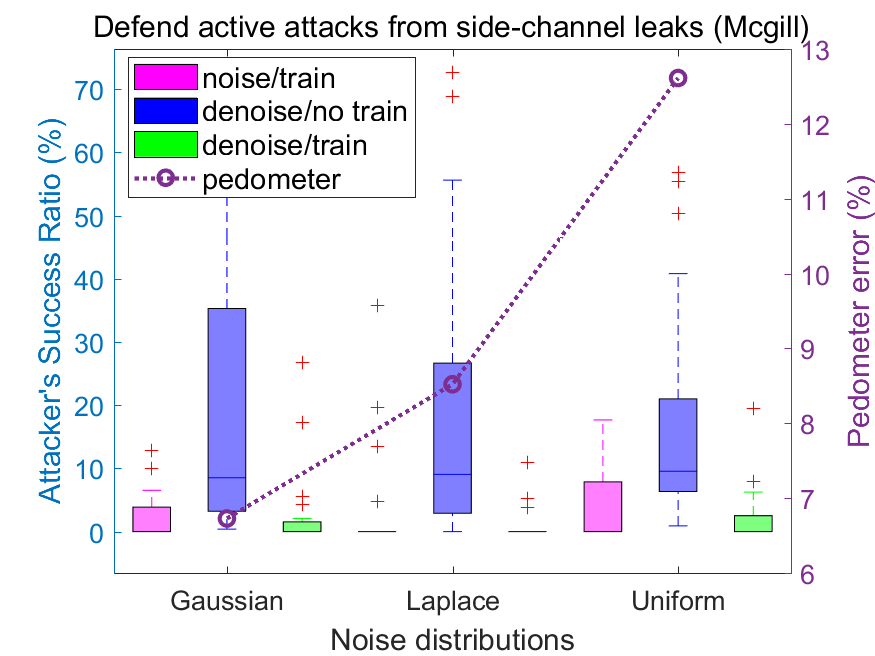}
                \vspace{-0.21in}
                \caption{}
\end{subfigure}
\hspace*{0.05in}
\begin{subfigure}[b]{0.28\textwidth}
                \includegraphics[width=1.05\textwidth]{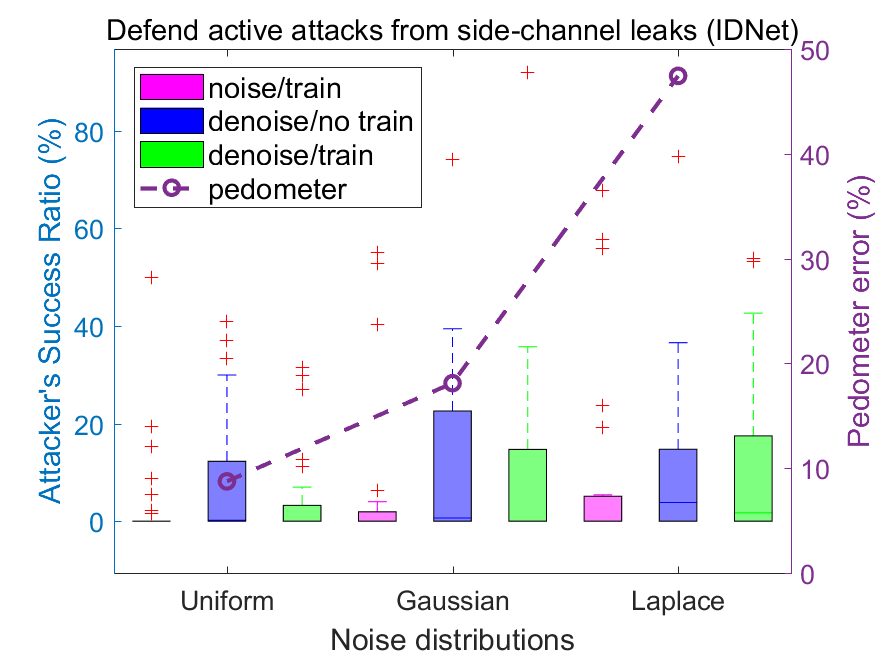}
                \vspace{-0.21in}
                \caption{}
\end{subfigure}
\hspace*{0.05in}
\begin{subfigure}[b]{0.28\textwidth}
                \includegraphics[width=1.05\textwidth]{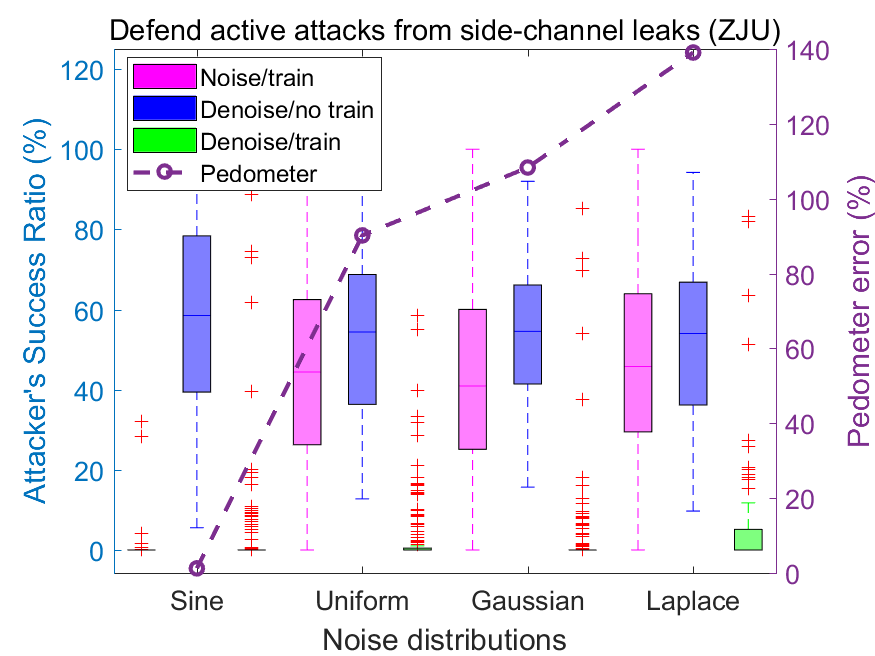}
                \vspace{-0.21in}
                \caption{}
\end{subfigure}%
\vspace*{-0.13in}
\caption{Defend active attacks through side-channel leaks (a) Mcgill; (b) IDNet; (c) ZJU.}
\label{side_channel_fig}
\vspace*{-0.11in}
\end{figure*}

Next, we evaluate the system robustness against \textit{active} attacks.
Sec.~\ref{sec:side_channel_attack} has shown that simple noise injection does not work well for obfuscation.
In addition to \emph{Gaussian} noise, we further evaluate \emph{Laplacian} and \emph{Uniform} noise with the standard deviation set to the original signal over a finite moving window. Laplacian noise is also used in differential privacy for mathematical tractability. We adopt the three types of noise to evaluate their properties regarding obfuscation and impact on usability. We choose a typical application of pedometer step counter to assess usability in the presence of noise. Fig.~\ref{side_channel_fig} shows the attacker's success ratio versus the pedometer error for different noise distributions. We alter the input in three ways. 1) \emph{noise/train}: noise samples are paired with genuine ones in the training set and labeled as negative. 2) \emph{denoise/no train}: attacker applies a state-of-the-art denoise technique called total variation proximity operators~\cite{prox-tv} on (1). The classifier takes no countermeasure. 3) \emph{denoise/train}: the classifier makes a successful prediction about the denoise scheme and labels the denoise pairs as negative for training.

Fig.~\ref{side_channel_fig}(a-b) indicate that the proposed mechanism is capable of defending against active attacks when the Siamese Network is supervised to learn the difference from the attack samples (noised or denoised). Learning the noised signals can drop the success rate from 50\% to less than 10\%.
However, without considering possible denoise from the attacker in the classifier, there is still around 20\% success rate even when the neural network has learned the noised signals.
Once denoise is considered in training, the attacker can no longer succeed. For usability, the noise distributions incur 7-13\% error for the step counter. IDNet gathered from a more vibrant environment has higher intra-class deviations. When the standard deviation of noise is set to as large as the original signal, the step counter is subject to a higher error rate.

An anomaly is ZJU in Fig.~\ref{side_channel_fig}(c), in which body sensors of low sampling rate are used. The additive noise has much higher frequency thus is bound to be filtered out by the neural networks. The error of step counter is almost doubled due to the local peaks of the noisy spikes being mistakenly recognized as gait cycles. Using random noise, we do not see much security improvement but a sharp usability degradation. Instead of noise distributions with high frequency, we further test a sinusoidal wave with a low frequency (identical to the gait signal with a much smaller amplitude). The sine wave is merged into the sensing signal and difficult to extract since the oscillating frequency is kept as a secret. On the other hand, the new frequency components are evident enough to be recognized by the neural network through training. As shown in Fig.~\ref{side_channel_fig}(c), the attacker's success ratio quickly drops to nearly zero for most of the $136$ individuals in ZJU. Our new finding suggests that random noise is not always a good solution to balance security and usability. The actual obfuscation should be considered with respect to the types of data source. For a better balance of security and usability, obfuscation with hidden regularity can be considered as a signature.

\begin{figure*}[!ht]
\centering
\includegraphics[width=1.65in]{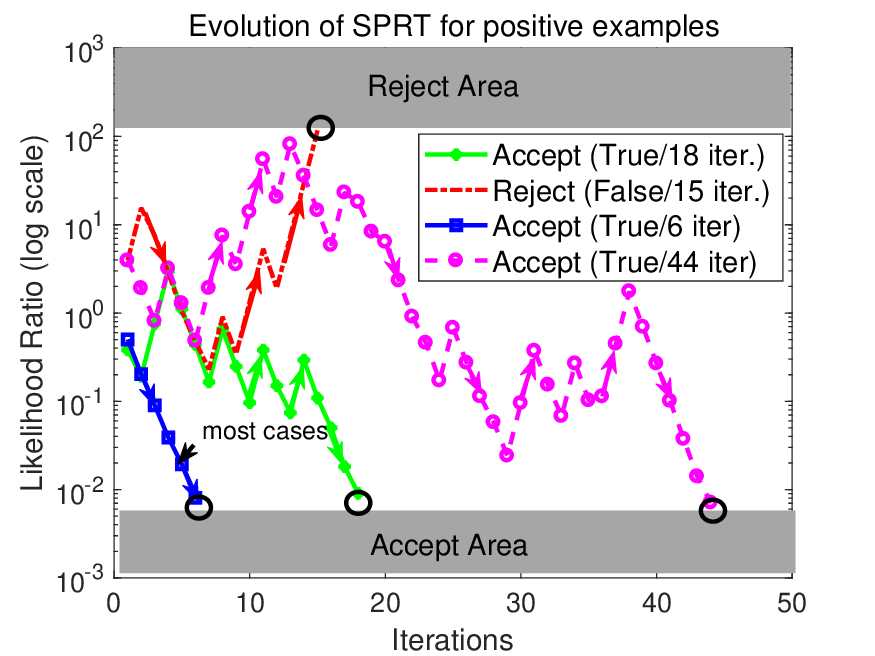}
\includegraphics[width=1.65in]{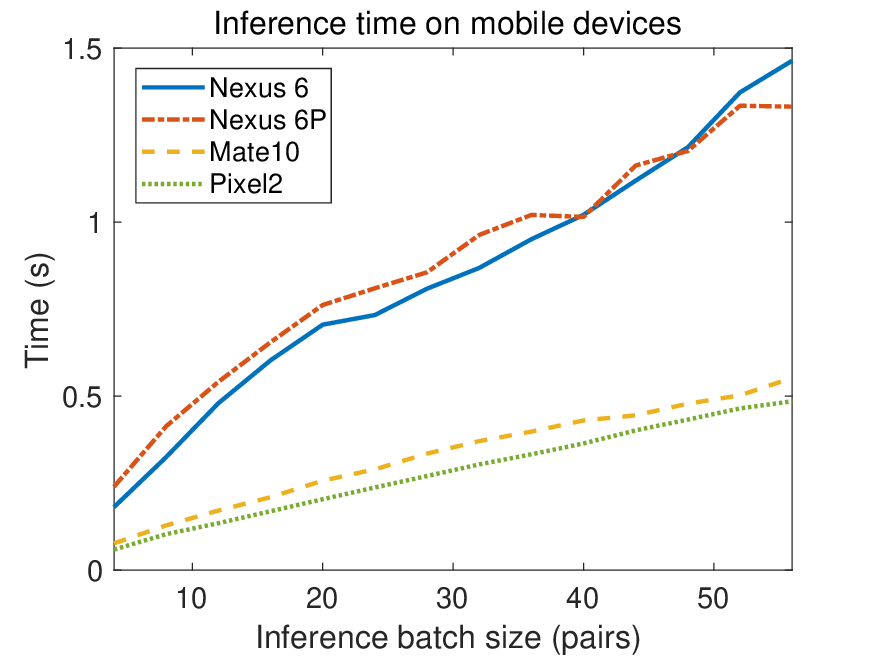}
\includegraphics[width=1.65in]{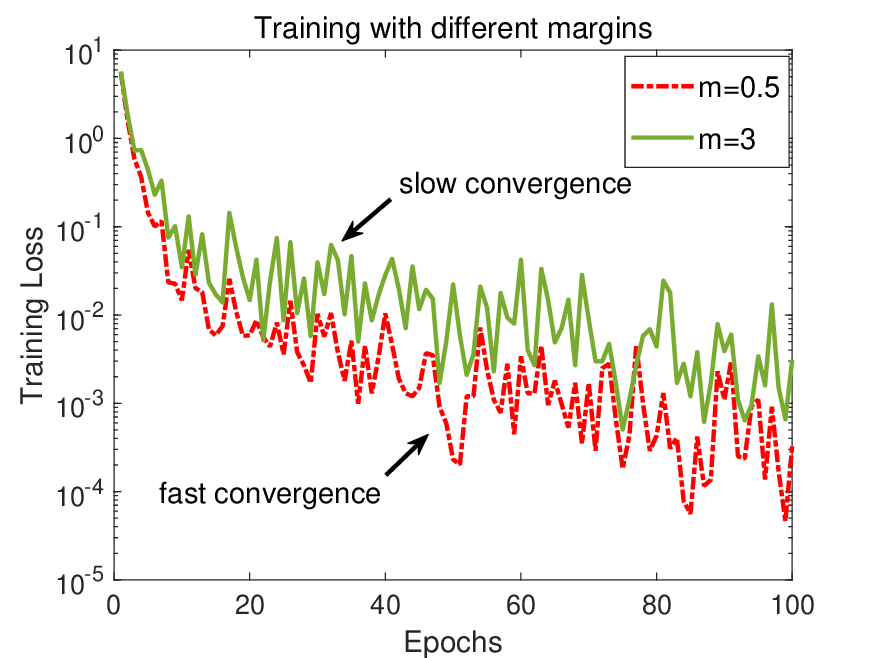}
\includegraphics[width=1.65in]{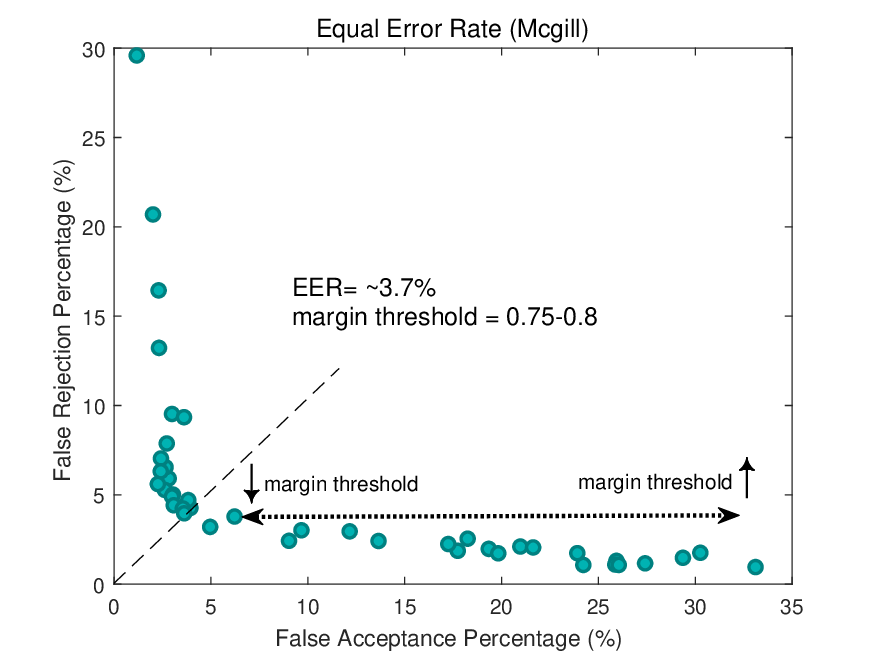} \\
{\small \hspace{0.1in} (a) \hspace{1.55in} (b) \hspace{1.55in}  (c) \hspace{1.55in} (d) }
\caption{System metrics (a) SPRT for positive samples (b) batched inference time on mobile (c) speed of training convergence with different margins (d) equal error rate (Mcgill). }
\label{miscellaneous_fig}
\vspace{-0.1in}
\end{figure*}


\vspace{-0.1in}
\subsection{Impacts from Accuracy Requirements and Margin}   \label{sec:miscellaneous}

This subsection evaluates the choice of two important parameters: the accuracy requirements in terms of false rejection/acceptance rate $\alpha,\beta$ and margin $m$ in the Siamese Network. $\alpha,\beta$ are inputs from the users. We set them to $0.01$ for demonstration. It means that we want the authentication to reach 99\% confidence about its decisions based on multiple observations. This is equivalent to $A=B=100$ in Eq. \eqref{sprt_eq}. For Eq. \eqref{proability}, the mean of distance is set to $m/2$ for balanced data and the variance are obtained on the training samples.

Fig.~\ref{miscellaneous_fig}(a) demonstrates the decision-making process. When the likelihood ratio hits the upper shaded area, the decision is to reject; otherwise, the decision is to accept. Normally, $5$-$6$ batch iterations are needed to reach a confident decision. To see more of how it evolves, we select some hard samples and mix them with random samples. Then the classifier is less confident based on the single batch, thus it progresses to the next iteration until a shaded region is hit. The process can be thought as a competition between the decisions to either accept or reject. If a majority of the new data indicates positive, the decision is inclined to accept though a few false ones may drag the curve towards the opposite direction en-route. As we can see, SPRT reduces authentication instability at a little cost of extended response time. Fig.~\ref{miscellaneous_fig}(b) shows time durations of making batched inference on mobile devices (from batch sizes of $4-56$). Since less parallel resources are available on the mobile platform (the CPU cores are fully utilized), the inference time increases almost linearly with the input batch size. The computation takes less than 1.5s for all the devices to process a single batch. In normal situations, reaching a confident decision of $5$-$6$ iterations takes about 6s and 1.5s on Nexus 6/6P and Pixel2/Mate10 respectively, which is quite acceptable for background processes.

The Siamese Network separates the positive and negative samples apart by \emph{margin} $m$. In our experiment, where the positive pair distance stays close to $0$, $m=0.5$ maps the negative pair distance to around $2$ and $m=3$ maps it to $4.5$. Intuitively, a small margin leads to higher error rate because dissimilar pairs are closer and possibly misclassified as similar pairs, or vice versa. A large margin makes it difficult to train the classifier in terms of slower convergence as shown in Fig.~\ref{miscellaneous_fig}(c). For balancing the rates between false acceptance and false rejection, the desirable margin is around $m=1.5$. We also demonstrate the margin threshold from $0.1$ to $3$ in Fig.~\ref{miscellaneous_fig}(d). If the distance is below the margin threshold, the test pair is similar or dissimilar otherwise. For Mcgill, our framework achieves an equal error rate (EER) around 96.3\% when the margin threshold is set to $\frac{1}{2}m$.



\vspace{-0.1in}
\subsection{Profile System Overhead}

\begin{figure}[t]
\centering
\hspace*{-0.3in}
\begin{subfigure}[b]{0.23\textwidth}
                \includegraphics[width=1.1\textwidth]{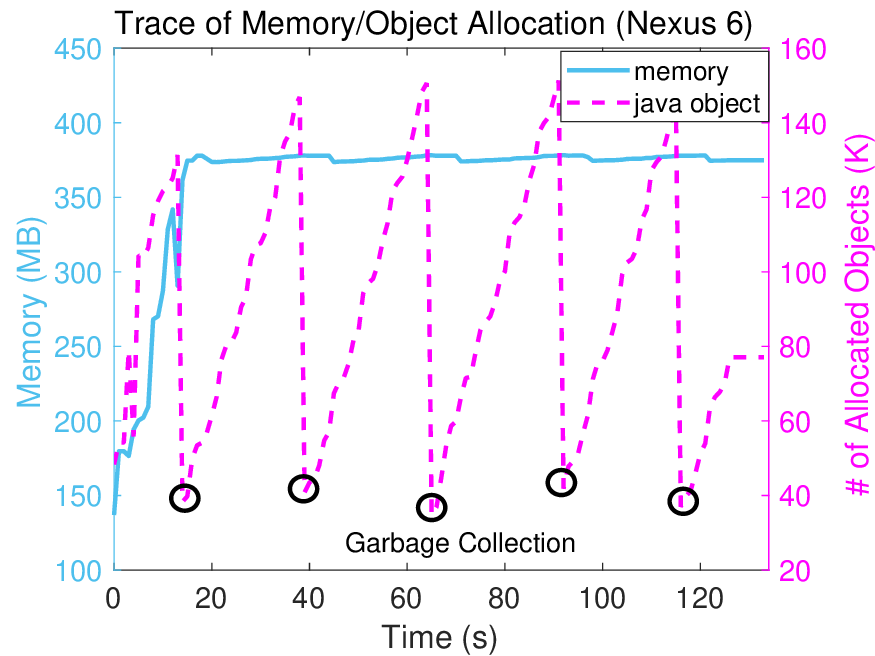}
\end{subfigure}
\hspace*{0.1in}
\begin{subfigure}[b]{0.23\textwidth}
                \includegraphics[width=1.1\textwidth]{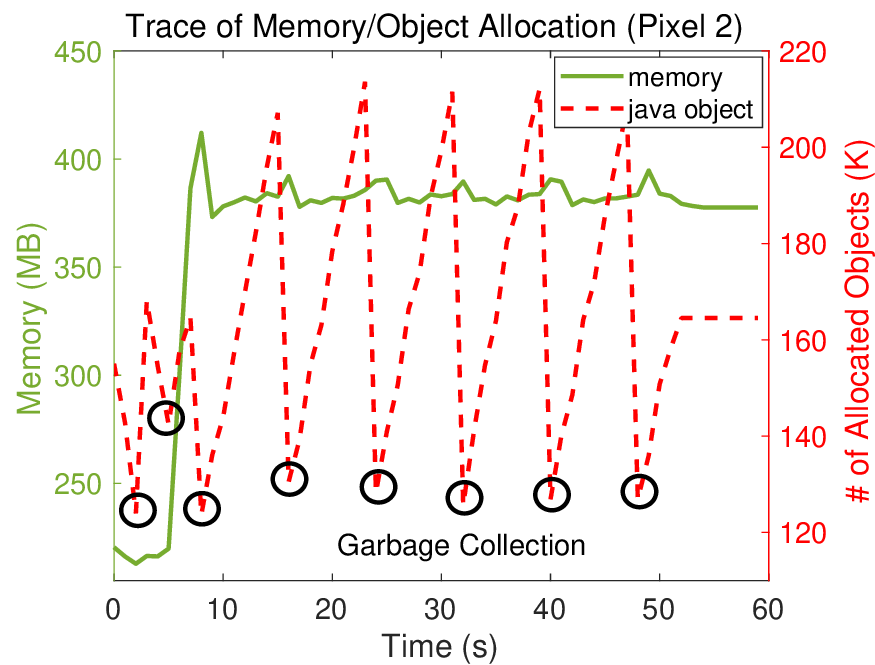}
\end{subfigure}%
\vspace*{-0.05in}
    {\small \\\hspace{0.01in} \textbf{(a) \hspace{1.55in}(b)}}
\vspace*{-0.1in}
\caption{Trace of memory/object allocation during mobile training (a) Nexus 6; (b) Pixel 2.}
\label{memory_fig}
\vspace*{-0.1in}
\end{figure}

\begin{figure}[ht!]
\centering
\hspace*{-0.3in}
\begin{subfigure}[b]{0.23\textwidth}
                \includegraphics[width=1.1\textwidth]{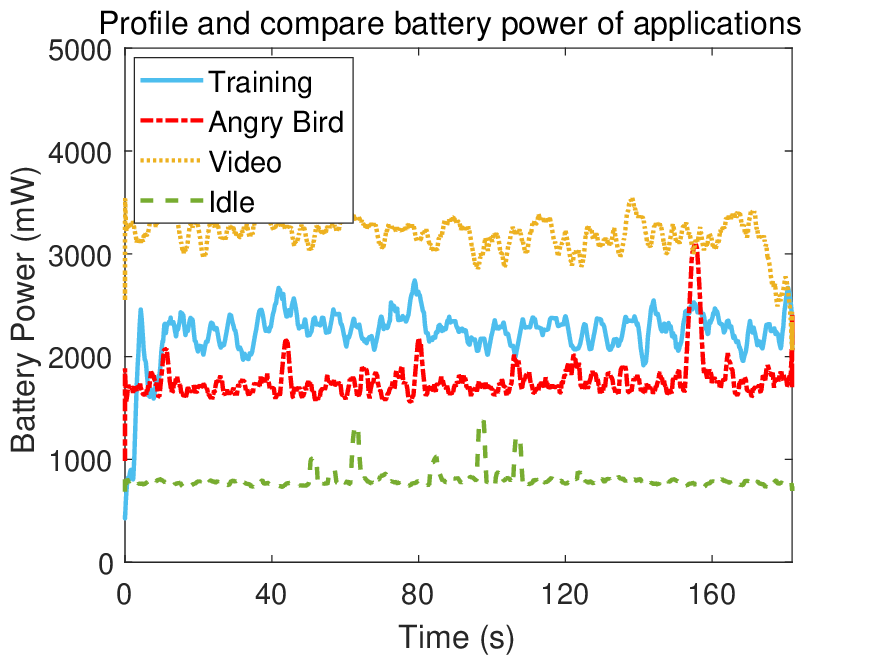}
\end{subfigure}
\hspace*{0.1in}
\begin{subfigure}[b]{0.23\textwidth}
        \includegraphics[width=1.1\textwidth]{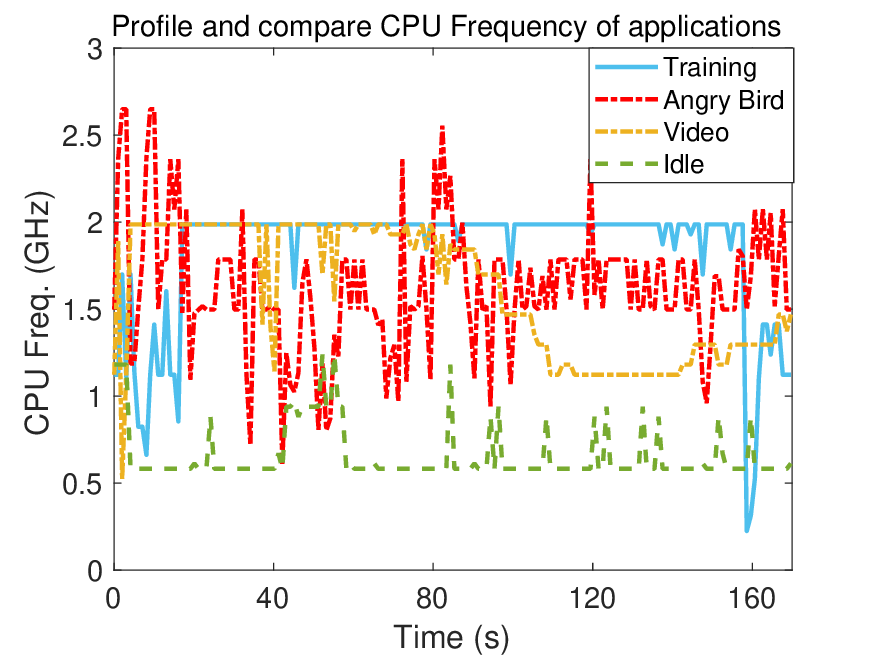}
\end{subfigure}
\vspace{-0.05in}
{\small \hspace{0.1in} (a) \hspace{1.55in} (b)}
\vspace{-0.05in}
\caption{Profiling battery power and CPU frequency of different applications}
\label{profile_fig}
\vspace*{-0.1in}
\end{figure}

\textbf{Memory.} We use the Android Profiler to measure the memory consumption of the app during training in Fig.~\ref{memory_fig}. To save space, we show the traces of Nexus 6 and Pixel 2 (the oldest and newest of our collection). Nexus 6 has a quad-core of $4 \times 2.7$ GHz. Pixel 2 features an octa-core with $4 \times 2.35$ GHz plus $4 \times 1.9$ GHz CPUs. Once the app starts, it loads the native code, training samples and network model into the mobile memory. Sample paring is conducted on the device at the beginning. Since DL4J is not optimized for the mobile environment, the native/code occupies about 130 MB. When training is initiated, new objects are allocated and once the app approaches the assigned memory limit, a garbage collection is triggered to release the objects, which could pause the app for a minimum amount of time (several ms). When multi-threads are enabled in DL4J with OpenBLAS, the training process enjoys much better performance with an octa-core processor on Pixel 2. Hence, we see a steeper line of object allocation on Pixel2, which completes the training by only half of the time with Nexus 6.


\textbf{Battery Power and CPU Frequency.} We profile the battery power using the Monsoon power monitor~\cite{monsoon} and CPU frequency by the Trepn Profiler\cite{trepn}. We measure the battery power and average CPU frequency of the 4 cores on Nexus 6 while (1) training, (2) playing angry bird, (3) watching an MP4 video in MX player, and (4) idling, in Fig.~\ref{profile_fig}. Training runs at 2.0 GHz set by the default governor and its battery power consumes at the level of 2000 mW, which consumes about 1\% total battery during 2.5 mins. Training introduces an additional 28\% energy overhead compared to angry bird, but consumes 25\% less energy compared to watching a video. The results suggest that training consumes more energy than mobile games but less intensive than watching videos. Since model update is less time-sensitive compared to interactive apps, it can be delegated as a background service and scheduled on-demand while the phone is charging or idling. The default CPU governor can be also adjusted adaptively to optimize performance and power consumption.

\vspace{-0.1in}
\section{Discussion}   \label{discussion:sec}

The primary goal of this paper is to meet the application requirements from authentication and explore whether the Android OS with the default settings can accommodate persistent workloads like training on consumer mobile devices (e.g., with the default \texttt{interactive} power management policy), in contrast to the conventional mobile workloads that are bursty in nature. Our implementation not only shows that it is feasible to execute training, but also reasonably fast using the multi-core CPUs (partially because we cannot run ultra deep models under the memory capacity). We keep the power management policy unchanged because the vendor-supplied drivers have heterogenous configurations regarding the CPU frequencies, proprietary task migration between the Big.LITTLE CPU clusters as well as the complex thermal behaviors. In the experiment, we notice some thermal throttling among the older generations (Nexus 6/6P), that the governor actively reduces frequency on the course of training or deactivates CPU cores at the big cluster. It leads to noticeable performance slowdown, but is an active measure to protect the CPU subsystem and battery from overheating.

To conserve energy, one direction on the OS level is to exploit co-running opportunities by scheduling learning with an appropriate foreground process~\cite{shen}, but at nontrivial performance trade-offs to slowdown the background learning process. Optimizations at the architecture level are more effective, such as improving the intra-layer~\cite{diannao,eyeriss}, cross-layer data locality~\cite{fuse-cnn,zcomp} and reducing data transfer via zero-value compression~\cite{dma-hpca}. These works are complementary to our research and we expect the designs to be integrated into specialized processors in the mobile architectures to carry out training tasks in the near future.

Other than authentication, the framework also provides the basis to launch federated learning tasks~\cite{fedavg,ipdps20}, a new computing paradigm for privacy-preserved collaboration among the mobile users, where on-device training is an essential element. To this end, we also expect that the proposed framework can provide a guideline to explore the design space for many federated applications looking forward.

\vspace{-0.1in}
\section{Conclusion}    \label{conclusion:sec}
This paper incorporates training on mobile devices and tackles the privacy and performance challenges for behavioral authentication. Empowered by deep metric learning, a comprehensive framework is designed to improve discriminative power and tackle side-channel leaks. Our extensive experiments demonstrate the security and robustness of the proposed design against intra-class variations and imposters that are out-of-distributions. We anticipate the presented system would offer insights and opportunities to enhance deep learning on mobile devices.

\section*{Acknowledgments}
This work was supported in part by the US NSF grant numbers CCF-1850045, CNS-1948131 and the State of Virginia Commonwealth Cyber Initiative (cyberinitiative.org).

\begin{IEEEbiography}
[{\includegraphics[width=1in,height=1.25in,clip,keepaspectratio]{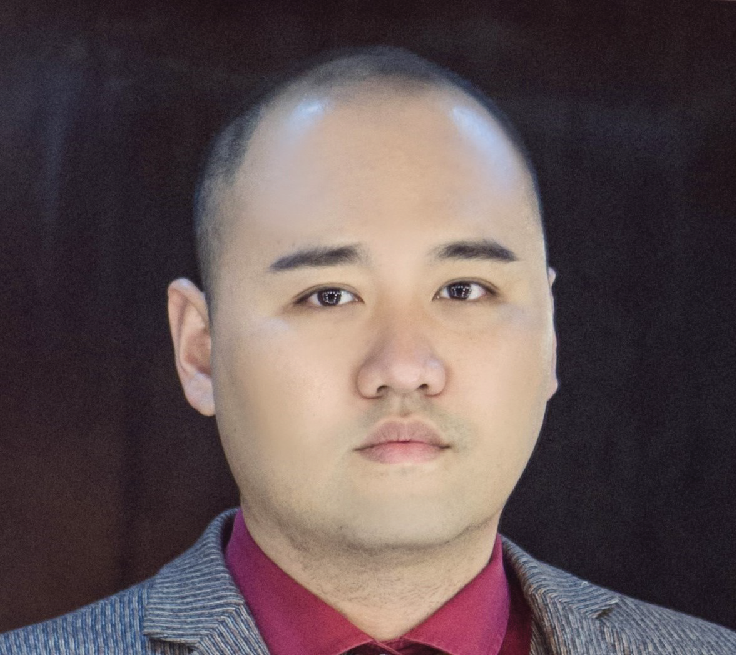}}]
{Cong Wang} received the B. Eng degree in Information Engineering from the Chinese University of Hong Kong in 2008, M.S. degree in Electrical Engineering from Columbia University in 2009, and Ph.D. in Computer and Electrical Engineering from at Stony Brook University, NY, in 2016. He is currently an Assistant Professor at the Computer Science department, Old Dominion University, Norfolk, VA. His research focuses on exploring algorithmic solutions to address security and privacy challenges in Mobile, Cloud Computing, IoT, Machine Learning and System. He is the recipient of Commonwealth Cyber Initiative Research and Innovation Award, ODU Richard Cheng Innovative Research Award and IEEE PERCOM Mark Weiser Best Paper Award in 2018.
\vspace{-0.1in}
\end{IEEEbiography}

\vspace{-0.1in}
\begin{IEEEbiography}
[{\includegraphics[width=1in,height=1.25in,clip,keepaspectratio]{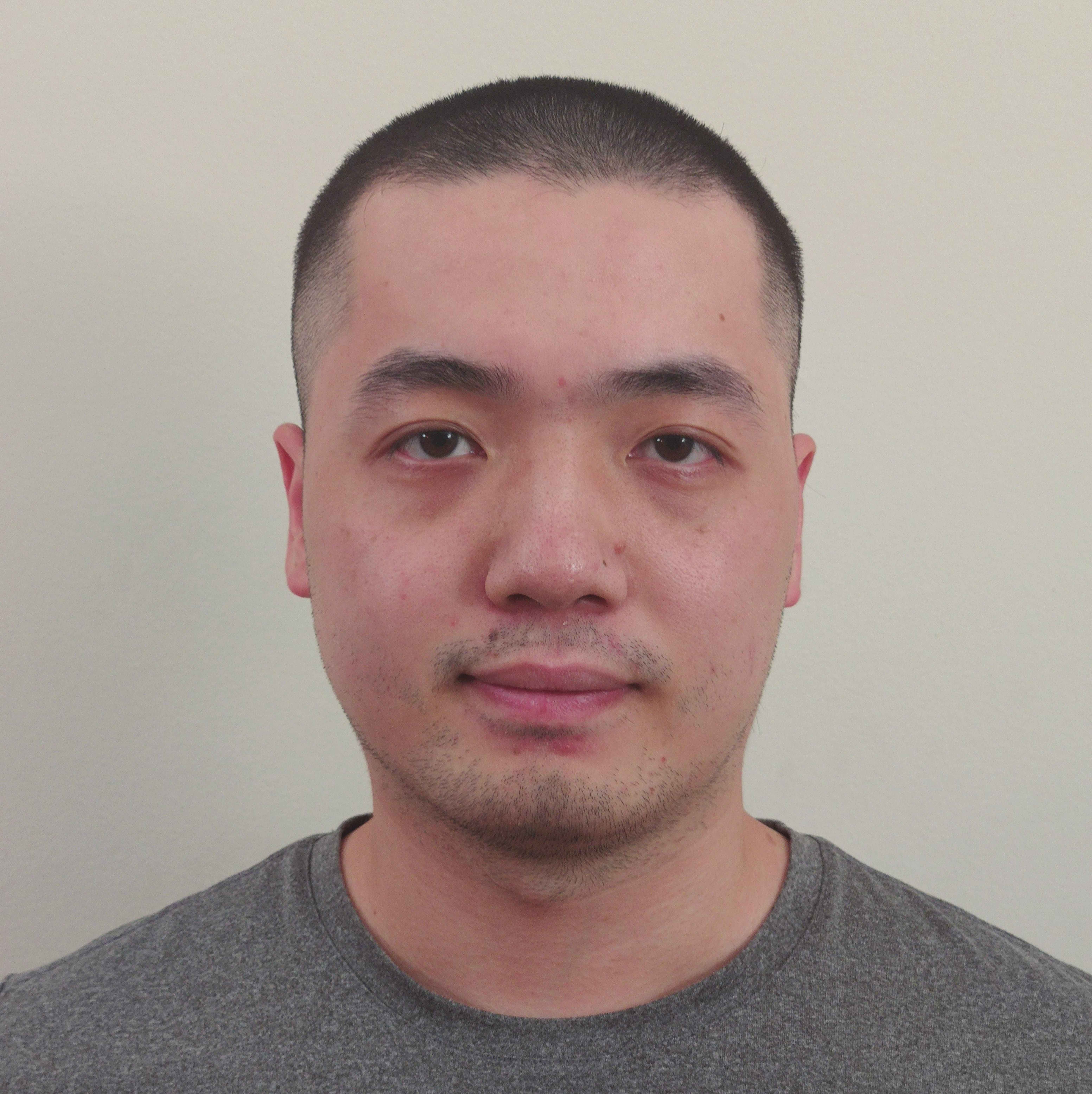}}]
{Yanru Xiao} received the B.S. degree in Computer Science from Central South University, China, 2017. He is currently pursuing PhD of Computer Science at Old Dominion University, VA. His research interests include artificial intelligence and security.
\vspace{-0.1in}
\end{IEEEbiography}

\vspace{-0.1in}
\begin{IEEEbiography}
[{\includegraphics[width=1in,height=1.25in,clip,keepaspectratio]{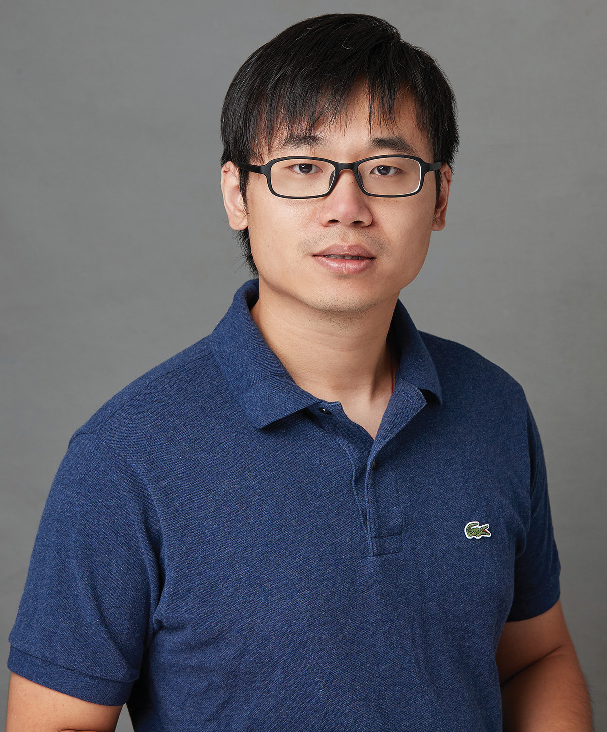}}]
{Xing Gao} received the Ph.D. degree in computer science from the College of William and Mary, Williamsburg, VA, USA, in 2018. He is an Assistant Professor in the Department of Computer and Information Sciences at the University of Delaware, Newark, DE, USA. His research interests include security, cloud computing, and mobile computing.

\vspace{-0.1in}
\end{IEEEbiography}

\vspace{-0.1in}
\begin{IEEEbiography}
[{\includegraphics[width=1in,height=1.25in,clip,keepaspectratio]{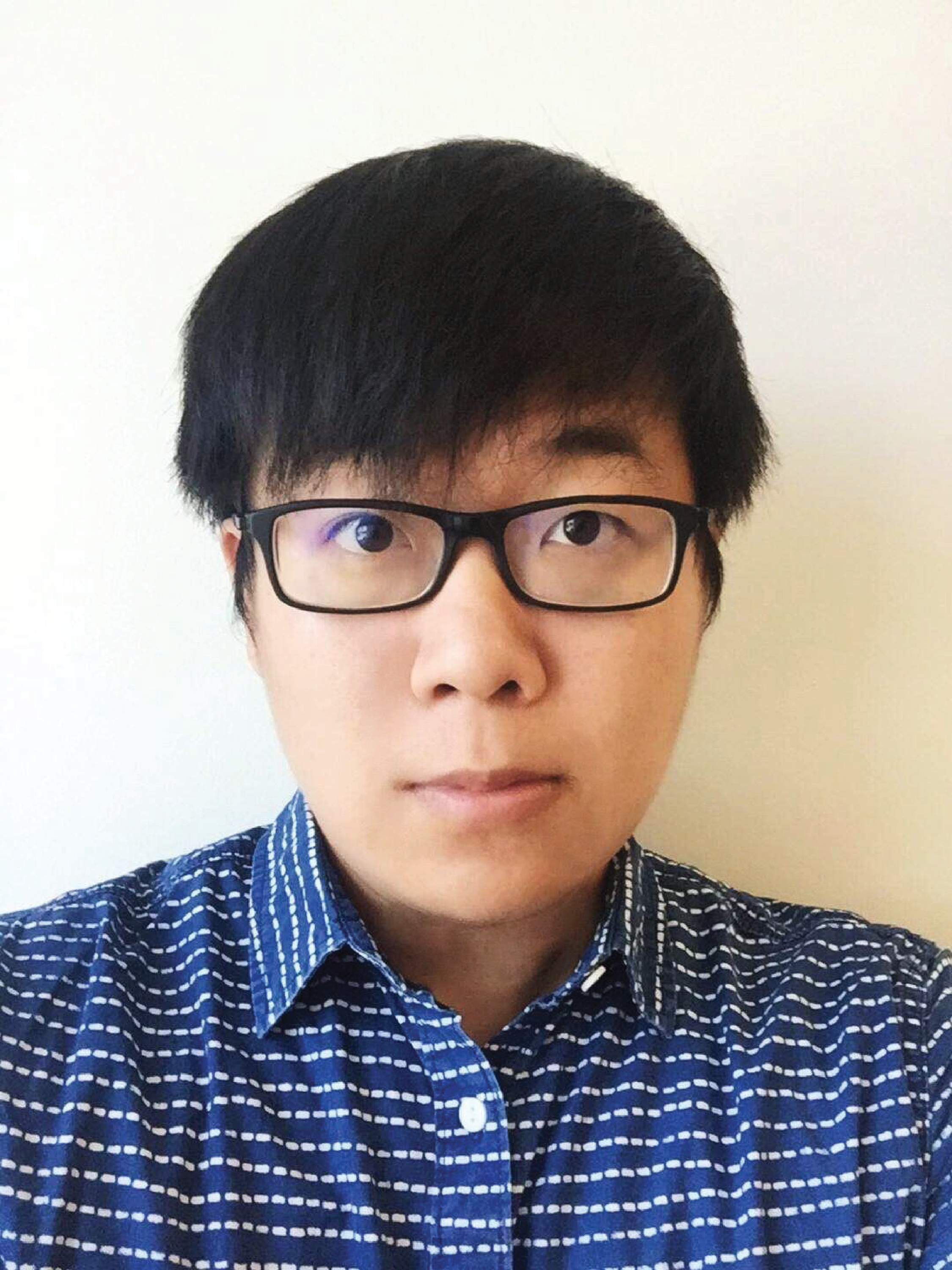}}]
{Li Li} is currently an Assistant Professor in ShenZhen Institutues of Advanced Technology, Chinese Academy of Sciences. He received his Ph.D. in Electrical and Computer Engineering from Ohio State University, Columbus, OH, USA in 2018. He received the M.S. degree in electrical and computer engineering from Ohio State University, Columbus, OH, USA, in 2014, and the B.S. degree in electrical engineering from Tianjin University, Tianjin, in 2011. His research interests include mobile computing and cloud computing.
\vspace{-0.1in}
\end{IEEEbiography}

\vspace{-0.1in}
\begin{IEEEbiography}
[{\includegraphics[width=1in,height=1.25in,clip,keepaspectratio]{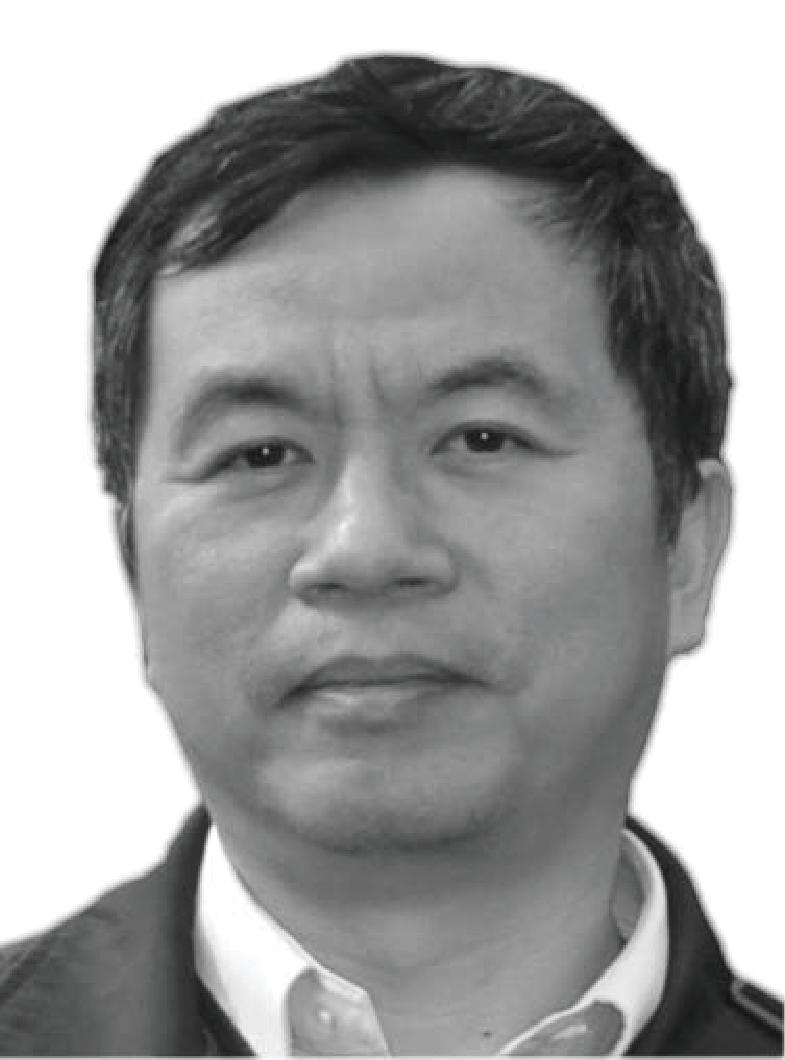}}]
{Jun Wang} received the PhD degree in computer science from McGill University. He is a senior researcher at Futurewei Technologies, in Santa Clara, CA, where he has been leading research projects on mobile computing, machine learning, and so on. His research interests include machine learning, mobile computing, compiler, computer architecture, among others. He is a member of the IEEE Computer Society.
\vspace{-0.1in}
\end{IEEEbiography}

\end{document}